%% file: main_arxiv.tex
\documentclass[a4paper,english,10pt,twocolumn]{article}
\usepackage{./style/arxiv}

\usepackage{amsmath,amssymb,amsfonts,amsthm}
\usepackage{dsfont} %
\usepackage{wrapfig}
\usepackage{mathtools}
\usepackage{xargs}
\usepackage{natbib}
\usepackage{algorithmic}
\usepackage{algorithm} %
\usepackage{bbold} %
\usepackage{hyperref}

\usepackage[markup=underlined, final]{changes} %

\usepackage{commands/shortcuts}
\newtheorem{theorem}{Theorem}
\newtheorem{assumption}{Assumption}

\makeatletter
\newcommand{\suppressmainTOC}{
  \let\addcontentsline@orig\addcontentsline
  \renewcommand{\addcontentsline}[3]{}
}
\newcommand{\restoreTOC}{
  \let\addcontentsline\addcontentsline@orig
}
\makeatother

\begin{document}

\title{MVICAD\textsuperscript{2}: Multi-View Independent Component Analysis with Delays and Dilations}

\date{}

\author{
    Ambroise Heurtebise\thanks{A. Heurtebise is with the University Paris-Saclay, Inria Saclay, and CEA (e-mail: ambroise.heurtebise@inria.fr).},
    Omar Chehab\thanks{O. Chehab is with the CREST at ENSAE (e-mail: emir.chehab@ensae.fr).},
    Pierre Ablin\thanks{P. Ablin is with Apple Paris.},
    and Alexandre Gramfort\thanks{A. Gramfort was with the University Paris-Saclay, Inria Saclay, and CEA. He is now with Meta Paris (e-mail: alexandre.gramfort@inria.fr).
    \newline
    \indent This work was supported by the ANR BrAIN (ANR-20-CHIA0016) grant.}
}

\suppressmainTOC

\maketitle

\begin{abstract}
    \input{content/01_abstract}
\end{abstract}

\section*{Introduction and related works}
\subsection*{Introduction}
\label{sec:introduction}

Machine learning techniques often encounter challenges when dealing with multiple views of the same underlying phenomenon.
\input{content/02_intro}

\input{content/03_method}

\input{content/04_regularization}
\section{Synthetic experiments}\label{sec:synthetic_experiments}
All the code is written in Python and is available on our \href{https://github.com/AmbroiseHeurtebise/mvicad}{GitHub repository}.
\input{content/05_synthetic_expes}

\input{content/06_real_data_expes}

\input{content/07_discussion}

\input{content/08_conclusion}

\input{content/09_acknowledgment}

\vskip 0.2in
\bibliography{references}

\newpage

\restoreTOC

\appendix
\onecolumn
\section*{Appendix}
The appendix is organized as follows.
\renewcommand{\contentsname}{}
\vspace{-1cm}
\tableofcontents
\newpage

\input{content/10_appendix}

\end{document}

%% file: content/01_abstract.tex
Machine learning techniques in multi-view settings face significant challenges, particularly when integrating heterogeneous data, aligning feature spaces, and managing view-specific biases.
These issues are prominent in neuroscience, where data from multiple subjects exposed to the same stimuli are analyzed to uncover brain activity dynamics. 
In magnetoencephalography (MEG), where signals are captured at the scalp level, estimating the brain's underlying sources is crucial, especially in group studies where sources are assumed to be similar for all subjects. 
Common methods, such as Multi-View Independent Component Analysis (MVICA), assume identical sources across subjects, but this assumption is often too restrictive due to individual variability and age-related changes.
Multi-View Independent Component Analysis with Delays (MVICAD) addresses this by allowing sources to differ up to a temporal delay.
However, temporal dilation effects, particularly in auditory stimuli, are common in brain dynamics, making the estimation of time delays alone insufficient. 
To address this, we propose Multi-View Independent Component Analysis with Delays and Dilations (MVICAD²), which allows sources to differ across subjects in both temporal delays and dilations. 
We present a model with identifiable sources, derive an approximation of its likelihood in closed form, and use regularization and optimization techniques to enhance performance.
Through simulations, we demonstrate that MVICAD² outperforms existing multi-view ICA methods.
We further validate its effectiveness using the Cam-CAN dataset, and showing how delays and dilations are related to aging.

%% file: content/02_intro.tex
These challenges include integrating heterogeneous data, aligning different feature spaces, and managing view-specific biases while maintaining consistency across views \cite{xu2013survey,sun2013survey,zhao2017multi}.
This is particularly evident in fields like neuroscience, where data from diverse subjects exposed to the same stimuli are analyzed to unveil the intricacies of brain activity.
In this context, a view corresponds to a subject.
For instance, a technique like magnetoencephalography (MEG) provides non-invasive recordings of magnetic fields produced by neurons in the brain.
Then, throughout MEG data analysis, signals captured at the scalp level can provide valuable insights into brain dynamics, by estimating sources of activity within the brain itself.
To draw general conclusions about these sources, group studies with hundreds of participants have been conducted with the objective of uncovering common sources for all subjects \cite{makeig2004mining, brookes2011investigating}.

When dealing with only one view, a common approach to estimate sources starts with Principal Component Analysis (PCA) to reduce the dimensionality of the observed signals.
Then, a single-view Independent Component Analysis (ICA) algorithm \cite{comon1994independent, hyvarinen2000independent, hyvarinen1999fast, ablin2018faster} is applied to extract independent sources.
Indeed, ICA relies on the assumption that components are independent and non-Gaussian, and it estimates the model's latent sources by maximizing the independence of the components according to some criteria.
However, the extension of ICA to a multi-view context is not straightforward.

State-of-the-art models in this domain such as Multi-View Independent Component Analysis (MVICA) \cite{richard2020modeling} and GroupICA \cite{calhoun2009review} assume that sources are shared and identical across views.
However, this assumption of exact source similarity across subjects is restrictive and may not always hold true.
Factors such as age or individual variability can lead to variations in the underlying sources, making them similar but not identical across different subjects \cite{haxby2011common, sabuncu2010function, cantlon2011cortical, curran2001effects, schmolesky2000degradation, finnigan2011erp}.

\citet{heurtebise2023multiview} relaxed this assumption with Multi-View Independent Component Analysis with Delays (MVICAD) by allowing sources to differ with a temporal delay across subjects.
However, biological studies have shown that temporal dilations\added{—that is, expansions or compressions of the time axis, as opposed to simple shifts in latency (delays)—are also}
common characteristics of neural responses, in particular for auditory tasks \cite{price2017age, walton2010timing, pichora2017older, alain1999age, schneider2010effects}. 

Building upon MVICAD, we introduce the Multi-View Independent Component Analysis with Delays and Dilations (MVICAD$^2$) algorithm.
MVICAD$^2$ allows sources to differ by a temporal delay \emph{and a dilation} across views.
This means that while all sources are shared across views, each source of each view may exhibit specific temporal variations, enhancing the model's flexibility and accuracy.
 
In Section \ref{sec:method}, we present our model, establish its identifiability, and derive an approximation of its likelihood which is cheap to compute.
We estimate the parameters of the model by maximizing this approximate-likelihood.
Section \ref{sec:regularization_optimization} focuses on regularization and optimization techniques used for this problem. 
Then, we evaluate our method using comprehensive simulations in Section \ref{sec:synthetic_experiments}.
Synthetic experiments with varying hyperparameters, such as the number of subjects, sources, or the noise level, demonstrate the superior performance of our method compared to existing multi-view ICA algorithms. 
Finally, in Section \ref{sec:real_data_experiments}, we evaluate MVICAD$^2$ on Cam-CAN, a large MEG population dataset \cite{taylor2017cambridge}. 
Our results reveal that the delays and dilations of neural responses recovered by our model correlate with age, in accordance with previous findings in the neuroscience literature \cite{price2017age}.

\subsection*{Related works}

Various ICA algorithms have been proposed to process data from multiple datasets at once.
PermICA \cite{richard2020modeling} is a naive approach to multi-view ICA.
It was designed as an initialization step for MVICA.
It consists in applying a single-view ICA algorithm, like FastICA \cite{hyvarinen1999fast}, to each view separately and then to reorder sources with the Hungarian algorithm \cite{kuhn1955hungarian} such that they are in the same order for all views.

Group ICA encompasses a range of ICA algorithms designed to process multiple views of the same data. This approach typically begins with the application of Principal Component Analysis (PCA) to each view independently. Next, the resulting sets are combined into a single dataset through either concatenation or multi-set Canonical Correlation Analysis (CCA)\cite{kettenring1971canonical, correa2010canonical}, possibly followed by an additional PCA step. An ICA algorithm is then applied to the consolidated set~\cite{varoquaux2009canica, tsatsishvili2015combining}. Concatenation can be performed spatially or temporally: temporal concatenation yields individual sources with a common mixing matrix, while spatial concatenation produces common sources with individual mixing matrices. Given our focus on shared sources, we compare our method with Group ICA using spatial concatenation\cite{calhoun2009review}.
Note that Goup ICA has shown inferior performance to MVICA on various neuroimaging tasks~\cite{richard2020modeling}.

MVICA~\cite{richard2020modeling} is our first reference algorithm.
It is designed to identify shared independent components across multiple datasets.
To do so, MVICA jointly estimates independent individual sources for each dataset and averages them to produce shared sources.
During optimization, the distance between individual and shared sources is minimized, which ensures that all views share approximately the same sources.
Unlike some other multi-view ICA approaches, MVICA's likelihood can be written in closed form and is then optimized with a quasi-Newton algorithm, using the gradient and an approximation of the Hessian of the likelihood.

MVICAD \cite{heurtebise2023multiview} is our second reference algorithm.
As in \cite{richard2020modeling}, it uses a quasi-Newton algorithm to estimate shared independent components across multiple datasets, but it allows individual sources to be time-delayed.
Optimization is done by alternating minimization with respect to delays and unmixing matrices (that will be defined later).

%% file: content/03_method.tex
\section{Method}\label{sec:method}

\paragraph*{Notation} In the following sections, lowercase letters are used for scalar values, lowercase bold letters are for vectors, capital letters denote matrices, and capital bold letters are used for multivariate signals.
A multivariate signal $\bS : \bbR \rightarrow \bbR^p$ can be evaluated at a time point $t \in \bbR$, as in $\bS(t) \in \bbR^p$, or at different time points $\bt \in \bbR^p$ simulatenously, as in $\bS(\bt)$ with slight abuse of notation. Additionally, applying a scalar-valued function $f$ to a matrix $S \in \bbR^{p \times n}$ means applying it to all the entries of $S$ and summing the results, as follows: $f(S) = \sum_{j, k} f(S_{jk})$.
Similarly, applying $f$ to a multivariate signal $\bS$ is defined by $f(\bS) = \sum_{j=1}^p \int_{t \in \bbR} f(\bS_j(t)) dt$, where $\bS_j$ is the $j$-th component of $\bS$.
The $\ell_2$ norm of a vector $\bs$ is $\|\bs\|$ and $\|S\|$ is the Frobenius norm of a matrix.
The norm of a multivariate signal is $\| \bS \| = \left( \sum_{j=1}^p \| \bS_j \|^2_{L^2} \right)^{\frac12}$, where $\| \cdot \|_{L^2}$ is the $L^2$ norm.
The set of integers $\{1, \dots, m\}$ is denoted $[\![m]\!]$.

\subsection{Model}

\begin{figure}[!t]
\centerline{\includegraphics[width=0.99\columnwidth]{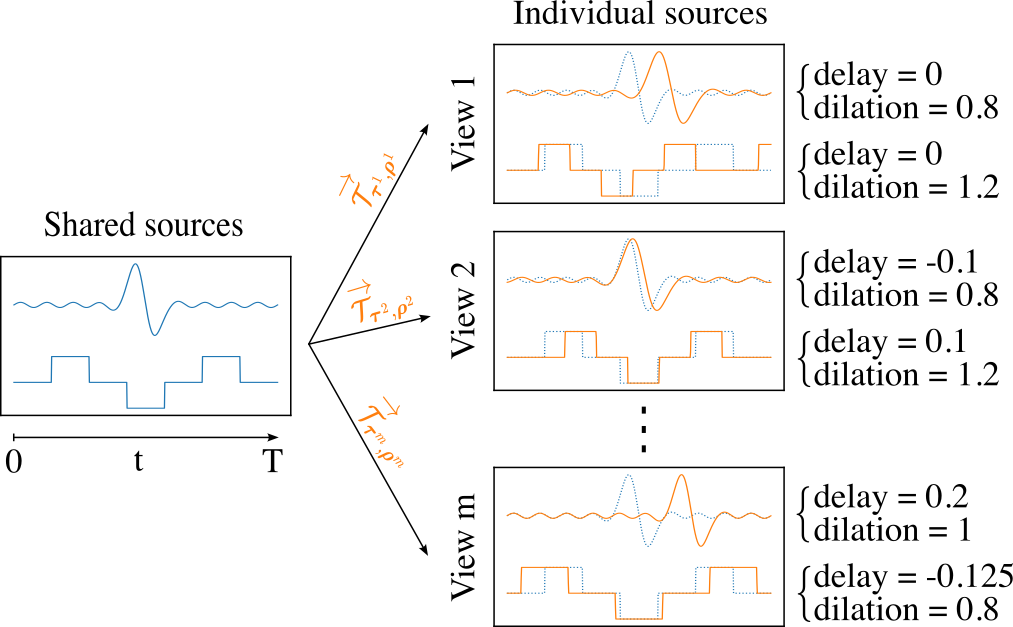}}
\caption{
\centering
Illustration of delays and dilations of sources.
There are $2$ components.
The multivariate signal on the left (in blue) represents the shared sources $\bS : [0, T] \rightarrow \bbR^2$.
Each individual source (in orange) is delayed and dilated, compared to its corresponding shared source.}
\label{fig:delays_and_dilations}
\end{figure}

\added{Consider $m$ views, $p$ sources, and time points $t \in \bbR$.}
We model the observed data $\bX^i(t) \in \bbR^p$ from view $i \in [\![m]\!]$ as a linear combination of  shared but temporally modified sources $\bS(t) \in \bbR^p$, plus noise.
Specifically, for each view $i$, we temporally modify the sources using delay parameters $\btau^i = (\tau^i_1, \ldots, \tau^i_p)$ valued in $[-\tau_{\max}, \tau_{\max}]$ and dilation parameters $\brho^i = (\rho^i_1, \ldots, \rho^i_p)$ valued in $[1/\rho_{\max}, \rho_{\max}]$.
These delays and dilations are applied using operators $\cD^1_{\btau^i}$ and $\cD^2_{\brho^i}$:
\begin{equation}
    \cD^1_{\btau^i}(\bS)(\bt) = \bS(\bt - \btau^i) \ \ \text{and} \ \ \cD^2_{\brho^i}(\bS)(\bt) = \bS(\brho^i \odot \bt) \enspace ,
\end{equation}
\added{where $\odot$ is the element-wise multiplication.}
Then, the composition $\overrightarrow{\cT}_{\btau^i, \brho^i} = \cD^2_{\brho^i} \circ \cD^1_{\btau^i}$ yields the temporal modification operator
\begin{equation}
    \overrightarrow{\cT}_{\btau^i, \brho^i}(\bS)(\bt) = \bS(\brho^i \odot (\bt - \btau^i))
\end{equation}
visualized in Figure~\ref{fig:delays_and_dilations}.

Using this operator, we can formally model the observed data as, $\forall i \in [\![ m ]\!]$, $\forall t \in \bbR$, 
\begin{equation}\label{eq:model}
    \bX^i(t) = A^i (\overrightarrow{\cT}_{\btau^i, \brho^i}(\bS)(t) + \bN^i(t)) \enspace ,
\end{equation}
where the $A^i \in \bbR^{p \times p}$ are view-specific mixing matrices, assumed to be full-rank, and the \added{$\bN^i : \bbR \rightarrow \bbR^p$}
are stochastic processes representing view-specific noise.
We assume that the $\bN^i$ are i.i.d. Gaussian distributed and independent across views with variance $\sigma^2$. In theory, $\sigma > 0$ should be estimated, yet, as shown in \cite{richard2020modeling} for MVICA, it has no critical impact on results and can be safely set to 1. Last, we assume the sources $\bS$ have components that are independent and i.i.d. across time, i.e. $\forall t_1, t_2 \in \bbR$, $t_1 \neq t_2$, $\bS(t_1)$ and $\bS(t_2)$ are i.i.d.

In the context of MEG data, the data of subject $i$ are usually reduced to only $p$ components with PCA, resulting in a multivariate signal $\bX^i : [0, T] \rightarrow \bbR^p$, where $T$ represents the duration of the experiment.
The hypothesis of shared sources up to time variability simply means that, after a common stimulus, brain activations from one subject to another have the same shape but differ up to delays and dilations.
Moreover, matrices $A^i \in \bbR^{p \times p}$ model the linear transformation between sources at the cortex level and observations at the scalp level.

\subsection{Identifiability}\label{sec:identifiability}

A model is said to be identifiable when it is described by one set of parameters, at most. Mathematically, this can be formulated as the injectivity of a mapping from parameters (here, mixing matrices and time delays and dilations) to the probability model of the observations.

\begin{assumption}\label{assump}
    \added{Assume that the sources $\bS$ are i.i.d. over time with distribution $\bbP_{\bS}$, which factorizes across components and is not degenerate. Let the noise terms $\bN^i$ be i.i.d. Gaussian with variance $\sigma^2$, mutually independent and also independent of the sources. Assume further that each mixing matrix $A^i$ is invertible, and that all components of $\brho^i$ are positive.}
\end{assumption}

\added{The following theorem gives a fundamental identifiability result, ensuring that recovering the sources is a well-posed problem under Assumption~\ref{assump}.}

\begin{theorem}[Identifiability]\label{thm:identifiability}
    \added{Let $(A^i, \btau^i, \brho^i, \bbP_{\bS}, \sigma)$ and $(A'^i, \btau'^i, \brho'^i, \bbP_{\bS'}, \sigma')$ be two sets that define the same statistical model in~\ref{eq:model}, for $i \in [\![ m ]\!]$. Under Assumption~\ref{assump}, the parameters $(A^i, \btau^i, \brho^i)$ and $(A'^i, \btau'^i, \brho'^i)$ are equal up to permutation and scaling indeterminacies, as specified in Equation~\eqref{app:eq:identifiability_conclusion} of Appendix~\ref{appendix:identifiability_proof}.}
\end{theorem}

The proof of the theorem is given in Appendix~\ref{appendix:identifiability_proof}.
    
\subsection{Approximate-likelihood}

As usual in ICA, we look for the inverse of the mixing matrices $A^i$, called unmixing matrices $W^i := (A^i)^{-1}$, since $W^i \bX^i$ approximately gives the individual sources $\overrightarrow{\cT}_{\btau^i, \brho^i}(\bS)$.
Given that we also want to estimate delays $\btau^i$ and dilations $\brho^i$, the likelihood of our model depends on three terms: $\cW = (W^1, \dots, W^m)$, $\btau = (\btau^1, \dots, \btau^m)$, and $\brho = (\brho^1, \dots, \brho^m)$. 
Note that we make a slight abuse of notations for $\btau, \brho \in \bbR^{m \times p}$, written in lowercase bold letters instead of capital letters.

Computing the likelihood of \eqref{eq:model} requires specifying the source distribution, typically assumed to be sub- or super-Gaussian.
For MEG data, neural sources are usually super-Gaussian due to their sparse and heavy-tailed nature \cite{bell1995information, hyvarinen1999fast}.
\added{We therefore fix a super-Gaussian density (e.g., logistic) for each source component.}
\added{As all components share the same distribution, }we now use $\bbP_{\bS}$ to denote both joint and marginal source densities.

For later use, we define a smoothed version $f$ of the source density $\bbP_{\bS}$ by convolving it with a Gaussian kernel $\cN(0_p, \frac{\sigma^2}{m} I_p)$, where $0_p$ is a vector of $p$ zeros and $I_p$ is the identity matrix of size $p$. The resulting function is given by
\begin{equation}\label{eq:function_f}
    f(s) := -\log \left( \int_u \bbP_{\bS}(s - u) \exp \left( -\frac{m}{2 \sigma^2}u^2 \right) du \right) \enspace .
\end{equation}

\added{Before introducing the likelihood, we also define the \emph{aligned estimated sources} for subject $i$ as}
\begin{equation}\label{eq:def_Y^i}
    \bY^i(t) := \overleftarrow{\cT}_{\btau^i, \brho^i}(W^i \bX^i)(t) \in \bbR^p \enspace, \quad t \in \bbR \enspace,
\end{equation}
\added{where $\overleftarrow{\cT}_{\btau^i, \brho^i}$ is the inverse of the time-warping operator $\overrightarrow{\cT}_{\btau^i, \brho^i}$. Intuitively, this operator removes the delays and dilations applied to the data.
We also define the \emph{average aligned sources} across all subjects as}
\begin{equation}\label{eq:def_Y_bar}
    \overline{\bY}(t) := \frac{1}{m} \sum_{i=1}^m \bY^i(t)
\end{equation}
\added{and, for later use, we denote by $\bY^i = (\bY^i(t))_{t \in \bbR}$ and $\overline{\bY} = (\overline{\bY}(t))_{t \in \bbR}$ the full aligned source time series for subject $i$ and the group average, respectively.}

\added{Using the above definitions, }we derive an approximation of the negative log-likelihood (NLL) of our model \added{over a given time interval $[0, T]$, expressed as}
\begin{align}\label{eq:approximate_NLL}
    \cL(\cW, \btau, \brho) &= -\sum_{i=1}^m \log|\det(W^i)| + \frac{1}{T} \int_{t=0}^T f \left( \overline{\bY}(t) \right) dt \nonumber \\
    &+ \frac{1}{T} \int_{t=0}^T \frac{1}{2 \sigma^2} \sum_{i=1}^m \left\| \bY^i(t) - \overline{\bY}(t) \right\|^2 dt \enspace .
\end{align}

\added{The derivation is detailed in Appendix~\ref{appendix:NLL_computations}.}
We emphasize that \eqref{eq:approximate_NLL} is an approximation---not the exact NLL---because of delays and dilations.
However, it tends towards the exact NLL when delays tend to 0, dilations tend to 1, and the following condition is verified: $\forall j \in [\![p]\!]$, $\forall i_1, i_2 \in [\![m]\!]$, $i_1 \neq i_2$, we have $\tau^{i_1}_j \neq \tau^{i_2}_j$ or $\rho^{i_1}_j \neq \rho^{i_2}_j$.
This last condition is almost sure if $\btau$ and $\brho$ are drawn uniformly.
See Appendix~\ref{appendix:NLL_computations} for details.

%% file: content/04_regularization.tex
\section{Regularization and optimization}\label{sec:regularization_optimization}

\subsection{Regularization terms}

\added{The final objective used in our method consists of the approximate-NLL defined in~\ref{eq:approximate_NLL}, supplemented with two regularization terms:}
\begin{equation}\label{eq:L_tilde}
\tilde{\cL} (\cW, \btau, \brho) 
= 
\cL (\cW, \btau, \brho) 
+ 
\lambda \cR_1 (\btau, \brho) 
+ 
\cR_2 (\cY, l) \enspace ,
\end{equation}
\added{where $\lambda \geq 0$ is a regularization scale, and $\cR_1$ and $\cR_2$ are defined below.}

\subsubsection{\texorpdfstring{Regularization \(\cR_1 (\btau, \brho)\)}{Regularization R1}}

\added{Without loss of generality, we can assume that the average delay vector $\btau$ has mean zero and the average dilation vector $\brho$ has mean one across subjects. 
Instead of enforcing these conditions as hard constraints—which would be difficult to handle in quasi-Newton optimization—we softly encourage them through a quadratic penalty. 
Specifically, $\cR_1 (\btau, \brho)$ penalizes deviations of the average delay $\overline{\tau}_j = \frac1m \sum_{i=1}^m \tau^i_j$ from zero, and of the average dilation $\overline{\rho}_j = \frac1m \sum_{i=1}^m \rho^i_j$ from one, for each source $j$. 
This term is multiplied by a regularization scale $\lambda \geq 0$, and its exact form is given in Appendix~\ref{appendix:regularization_R1}. 
An empirical study in the same appendix shows that performance remains stable for $\lambda$ in the range $[10^{-7}, 10^3]$, justifying the choice $\lambda=1$ in all experiments.}

\subsubsection{\texorpdfstring{Regularization \(\cR_2 (\cY, l)\)}{Regularization R2}}

Unfortunately, the approximate-NLL in~\eqref{eq:approximate_NLL} is non-convex, as illustrated by the indeterminacies of our model.
For example, in the presence of a periodic signal, delays that are equal to any multiple of the signal's period (within the authorized bounds) correspond to local minima, even in the presence of the regularization term $\cR_1$.
To mitigate this, we add another regularization term $\cR_2$ that captures information about the sources' envelope\added{ by comparing smoothed source magnitudes to their average across subjects}.

Using the definition of $\bY^i$, we define the collection of fully aligned estimated sources as $\cY = (\bY^1, \dots, \bY^m)$.
\added{Smoothing is performed using the operator $\cS_l$, defined as a convolution with a uniform kernel of length $l \in \bbN$, i.e., a moving average over $l$ consecutive time points. The regularization term reads:}
\begin{equation}\label{eq:regularization_term_2}
    \cR_2(\cY, l) = \frac{1}{2 \sigma^2} \sum_{i=1}^m \left\| \cS_l( \lvert \bY^i \rvert) - \overline{\bS} \right\|^2 \enspace ,
\end{equation}
where $\overline{\bS} = \frac1m \sum_{i=1}^m \cS_l( \lvert \bY^i \rvert)$\added{ is the average smoothed source magnitude across subjects, and $| \cdot |$ denotes the absolute value function}.
\added{Appendix~\ref{appendix:filter_length_l} highlights the importance of the regularization term $\cR_2$, and shows that the choice of $l$ is not critical, provided it is not excessively large. We use $l=3$ by default.}

\subsection{Refining parameters' scales}\label{sec:refining_parameters_scales}

We chose the limited-memory BFGS (L-BFGS) \cite{liu1989limited} algorithm to optimize \eqref{eq:L_tilde} with respect to the parameters $\cW$, $\btau$, and $\brho$.
Since $\btau$ and $\brho$ are bounded, we opted for a version of L-BFGS that allows parameters to be constrained, namely L-BFGS-B \cite{byrd1995limited}.
This algorithm requires the gradient $G$ and an approximation of the Hessian $\cH$ of the loss, which is easily done with automatic differentiation techniques as implemented by the JAX library \cite{jax2025}.

L-BFGS-B finds the descent direction by refining the gradient of \eqref{eq:L_tilde} with curvature information derived from previous steps.
This makes the algorithm robust to differences in parameters' scales.
However, in the first iteration, curvature information, represented by the Hessian of the loss, is assumed to be the identity.
Thus, having parameters $\cW$, $\btau$, and $\brho$ of different scales can be detrimental to the optimization.
To address this problem, we correct scale dissimilarities at the beginning of the algorithm.

\subsubsection{Scale of delays VS dilations}
First, we align delays' scale with dilations' scale.
To do so, for a given duration $T > 0$, we measure how much varying some delays $\tilde{\btau} \in \bbR^p$ around $\mathbf{0}$ and dilations $\tilde{\brho} \in \bbR^p$ around $\mathbf{1}$ affects the quantity
\begin{equation}\label{eq:loss_Lambda}
    \int_0^T  \left\| \overrightarrow{\cT}_{\tilde{\btau}, \tilde{\brho}} (\overline{\bY})(t) - \overline{\bY}(t) \right\|^2 dt \enspace .
\end{equation}
Of course, \eqref{eq:loss_Lambda} equals $0$ when $\tilde{\btau} = \mathbf{0}$ and $\tilde{\brho} = \mathbf{1}$.
Also, one can observe that \eqref{eq:loss_Lambda} looks like the third term of \eqref{eq:approximate_NLL}, up to a scaling factor and without the sum over views.
Indeed, $\overrightarrow{\cT}_{\tilde{\btau}, \tilde{\brho}} (\overline{\bY})$ is a delayed and dilated version of the shared sources $\overline{\bY}$, so it looks like the individual sources $\bY^i$ for specific $\tilde{\btau}$ and $\tilde{\brho}$.
Calculating the Hessian of \eqref{eq:loss_Lambda} with respect to $\tilde{\btau}$ and $\tilde{\brho}$ then allows to change delays' scale, so that varying equally delays around $\mathbf{0}$ or dilations around $\mathbf{1}$ affects equally \eqref{eq:loss_Lambda}.

\added{Ultimately, we obtain a scalar $\Lambda_j$ for each source $j$, which is then used to rescale the delays $\tau^i_j$ across all subjects. We denote by $\Lambda^1 := \{\Lambda_1, \ldots, \Lambda_p\}$ the resulting vector of scaling factors. Further details on the computation of $\Lambda_j$ are provided in Appendix~\ref{appendix:parameters_scales_1}.}

\subsubsection{Scale of temporal parameters VS unmixing matrices}\label{sssec:Lambda_2}
Second, we multiply the scale of both time parameters (delays and dilations) by a hyperparameter $\Lambda^2 \in \bbR$.
Increasing this hyperparameter allows to give more importance to the optimization of delays and dilations, compared to the optimization of unmixing matrices.
In practice, we find that $\Lambda^2$ plays a crucial role and must be set to a sufficiently high value. \added{However, as detailed in Appendix~\ref{appendix:parameters_scales_2}, performance remains stable across a broad range of values, making precise tuning unnecessary}.

\begin{figure*}[!t]
\centerline{\includegraphics[width=2.03\columnwidth]{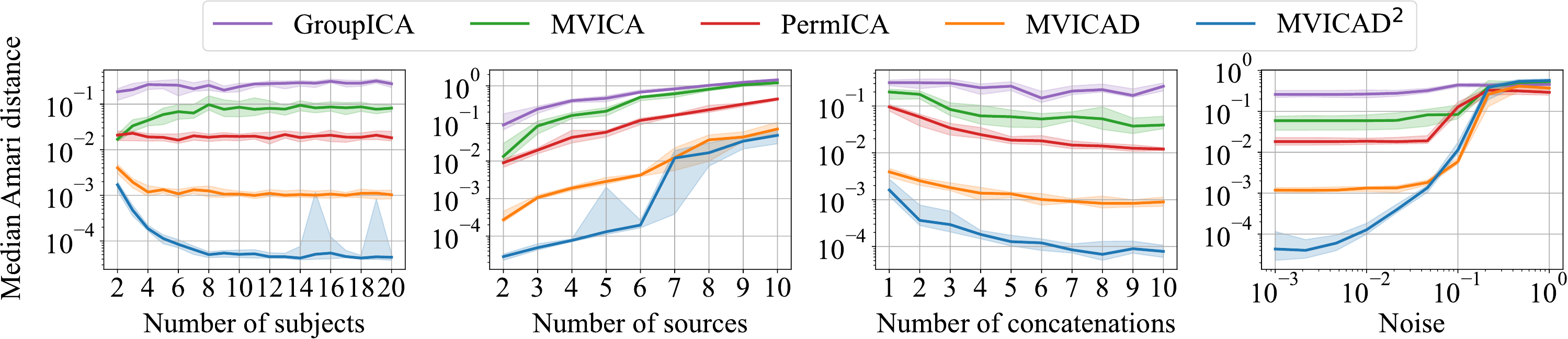}}
\caption{
\centering
Amari distance of multiple methods with respect to the number of subjects (upper left), number of sources (upper right), number of concatenations (lower left), and noise level (lower right).
We used $30$ different seeds and plotted the median result over all seeds.}
\label{fig:synthetic_expes_amari_distance}
\end{figure*}

\begin{figure*}[!t]
\centerline{\includegraphics[width=2.03\columnwidth]{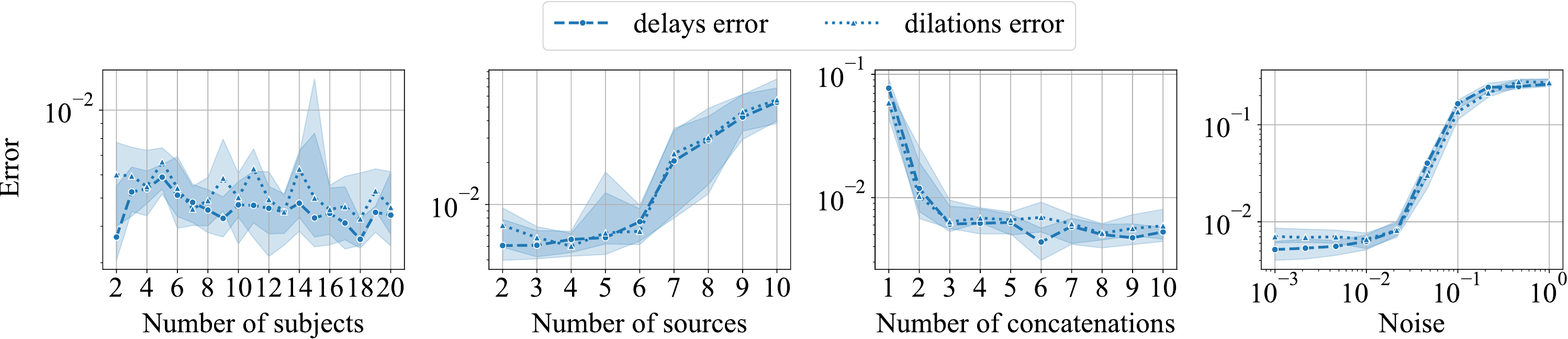}}
\caption{
\centering
Delays error $d_{\text{delays}}$ and dilations error $d_{\text{dilations}}$ of our method with respect to the number of subjects (upper left), number of sources (upper right), number of concatenations (lower left), and noise level (lower right).
We used $30$ different seeds.}
\label{fig:synthetic_expes_time_params_error}
\end{figure*}

\subsection{Initialization}\label{sec:initialization}

A naive way to initialize the parameters of our algorithm would be to start from random unmixing matrices, delays equal to $0$, and dilations equal to $1$.
But to avoid starting far from the true parameters and eventually finding a bad local minimum, we initialize separately the unmixing matrices of each view, and then delays and dilations.
A simple single-view ICA algorithm suffices to initialize the unmixing matrices separately but may estimate sources in different orders, from one view to another, because of the permutation indeterminacy inherent to ICA.
We use Picard \cite{ablin2018faster} to perform single-view ICA.
To find the permutations that put sources of all views in the same order, MVICA \cite{richard2020modeling} correlates individual sources with sources of a reference view, and then uses the Hungarian algorithm \cite{tichavsky2004optimal}.
In our case, due to delays, one could consider a cross-correlation followed by the Hungarian algorithm.
However, due to dilations, this is not sufficient.
Therefore, we propose to search initial delay and dilation values over a grid of candidate values and use correlation of delayed and dilated sources with the sources of the reference view.
Algorithm~\ref{algo:initialization} (in Appendix~\ref{appendix:initialization})
describes how the initialization works and Algorithm~\ref{algo:main} (in Appendix~\ref{appendix:main_algorithm})
summarizes the entire pipeline.

\subsection{Interpolation and boundary conditions}
\label{ssec:interpolation_and_boundary_conditions}

In practice, MVICAD$^2$ takes as input a collection $\cX = (\bX^1, \dots, \bX^m)$ of multivariate signals that contain a fixed number of time instants, typically defined over an evenly sampled grid.
When dealing with operator $\overleftarrow{\cT}$, changing the time axis with a continuous delay or dilation means evaluating the signal of the initial grid.
To address this problem, we use linear interpolation, that is also amenable to automatic differentiation.

Furthermore, multivariate signals have a fixed duration $T > 0$.
Delaying and dilating signals sometimes requires to evaluate them at $t < 0$ or $t > T$.
To address these cases, we use cyclic boundary conditions.
In other words, we define sources $\bS$ outside of $[0, T]$ by saying that $\bS$ is a periodic multivariate signal with period $T$.
Although sometimes unsuitable for real signals, having boundary conditions does not produce abrupt discontinuities at the edges of the signal if the sources start and finish at the same level.
\added{This condition is typically satisfied in MEG data due to standard high-pass filtering, which removes slow drifts and ensures that the signal begins and ends at similar levels. For other types of data, this property can be enforced by applying a windowing function, such as a Hamming window, which smoothly attenuates the signal amplitude at the edges.}

\subsection{Differences with MVICAD}\label{ssec:differences_with_MVICAD}

We now clarify the key differences between MVICAD~\cite{heurtebise2023multiview} and the MVICAD\textsuperscript{2} framework presented in this work.
First, MVICAD models only delays, whereas MVICAD\textsuperscript{2} jointly estimates both delays and dilations.
Second, MVICAD\textsuperscript{2} estimates continuous-valued parameters, \added{allowing for more precise modeling, }while MVICAD relied on discretized estimates.
Third, during delay estimation, MVICAD simplified optimization by removing the function $f$ from the loss, whereas MVICAD\textsuperscript{2} optimizes the full objective function without such approximations.
Finally, MVICAD minimized the loss sequentially across views, while MVICAD\textsuperscript{2} performs joint optimization over all parameters \added{using JAX-based automatic differentiation and an L-BFGS-B optimizer.
These algorithmic differences result in consistently better performance across experiments.}

%% file: content/05_synthetic_expes.tex
In our experiments, we compare our method to the algorithms PermICA \cite{richard2020modeling}, GroupICA \cite{calhoun2009review}, MVICA \cite{richard2020modeling}, and MVICAD \cite{heurtebise2023multiview}.
Comparisons with other non-ICA methods such as the Shared Response Model (SRM) \cite{chen2015reduced} and BCorrCA \cite{kamronn2015multiview} were already made in \cite{richard2020modeling}.

\subsection{Synthetic data}

\begin{wrapfigure}{r}{0.435\columnwidth}
    \vspace{-25pt}
    \centering
    \includegraphics[width=0.435\columnwidth]{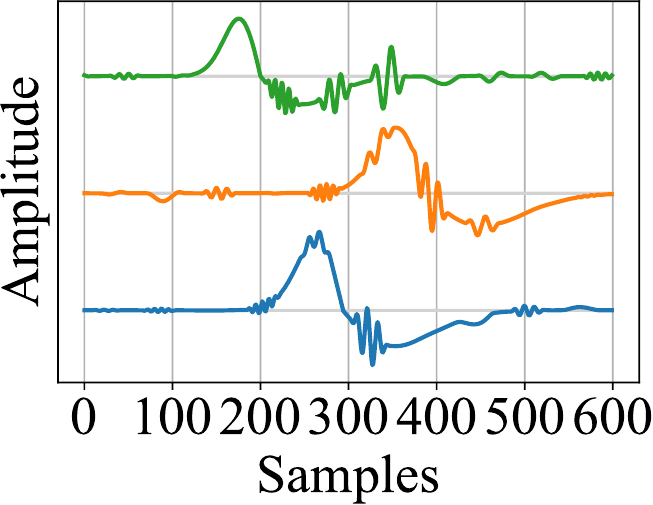}
    \vspace{-5pt}
    \caption{
    \centering
    Example of synthetic sources. We used $p=3$ and $n=600$.}
    \vspace{-5pt}
    \label{fig:synthetic_sources}
\end{wrapfigure}
We generate synthetic data following our model \eqref{eq:model}.
Except for figures that precisely aim to vary these parameters, we choose the number of views $m=5$, the number of sources $p=3$, the number of samples $n=600$, the maximum delay $\tau_{\max} = 0.05$, and the maximum dilation $\rho_{\max} = 1.15$.
Matrices $A^i$ are randomly drawn with a Gaussian i.i.d. distribution, vectors $\btau^i$ (resp. $\brho^i$) are uniformly drawn in $[-\tau_{\max}, \tau_{\max}]$ (resp. $[1/\rho_{\max}, \rho_{\max}]$), and all the elements of the noise matrices $\bN^i$ are i.i.d. and follow the distribution $\cN(0, \sigma^2)$ with $\sigma$ set to 1.

To generate the sources, we use the function $x \mapsto -x \exp(-x^2)$.
This function has a positive peak to the left of 0 and a negative peak to the right of 0 that we halve.
The heights, means and variances of the peaks are different from one source to another. 
Then, we add small features by decomposing the time axis into 10 different bins and generating sin waves of different frequencies in each bin, before applying a Hamming window in each segment.
Typical sources are displayed in Figure \ref{fig:synthetic_sources}.

Finally, to increase the statistical performance of the method, we generate these kind of sources $n_{\text{concat}} = 5$ times and temporally concatenate them, so the source matrix has the shape ${p \times n_{\text{concat}}n}$.
We also sample $n_{\text{concat}}$ times the noise matrices, so that the $N^i$ have the same shape as $S$.

\subsection{Metrics}

\added{The performance of ICA algorithms is commonly assessed using the Amari distance} \cite{moreau1998self}\added{, which compares estimated unmixing matrices $W^i$ to true mixing matrices $A^i$. This metric is invariant to scaling and permutations, and increases as $W^i A^i$ deviates from a scale-permutation matrix.}

To quantify the error on delays, we also introduce the quantity $d_{\text{delays}}$ that takes as inputs the true delays $\btau_{\text{true}} \in \bbR^{m \times p}$ and the estimated delays $\btau \in \bbR^{m \times p}$.
True and estimated delays are mapped from $[-\tau_{\max}, \tau_{\max}]$ onto $\left[-\frac12, \frac12 \right]$ to make them independent of $\tau_{\max}$, and they are centered around $0_p$ because they can only be estimated up to a global delay.
In doing so, we obtain $\hat{\btau}_{\text{true}}$ and $\hat{\btau}$.
We define $d_{\text{delays}}(\btau_{\text{true}}, \btau) \in [0, 2[$ as the mean of $| \hat{\btau}_{\text{true}} - \hat{\btau} | \in \bbR^{m \times p}$.

Following the same procedure, we define the quantity $d_{\text{dilations}}$.

\subsection{Synthetic results}\label{ssec:synthetic_results}

We now present results obtained from synthetic experiments.
Figure \ref{fig:synthetic_expes_amari_distance} shows how well the unmixing matrices are estimated. Specifically, it tracks how the Amari distance of several methods behaves when changing the numbers of subjects, sources, and concatenations, and when we vary the noise level.
Results were obtained from 30 different seeds.
We observe that our method outperforms the other algorithms by an order of magnitude, except when the noise level or the number of sources is too big.
Note that, for the plot with respect to the number of sources, we used 10 views instead of 5.
However, we can hypothesize that the increase of Amari distance when the number of sources exceeds 6 partially comes from our source generation.
Indeed, synthetic sources are generated in a similar way and the possibility to delay and dilate them can make source separation more tedious, as the number of sources increases.

Figure~\ref{fig:synthetic_expes_time_params_error} reports the estimation errors for delays and dilations in the same four settings as Figure~\ref{fig:synthetic_expes_amari_distance}. We observe that these errors closely follow the trend of the unmixing matrices error, suggesting that source separation improves as temporal parameter estimation becomes more accurate.

\begin{figure}[!t]
\centerline{\includegraphics[width=0.85\columnwidth]{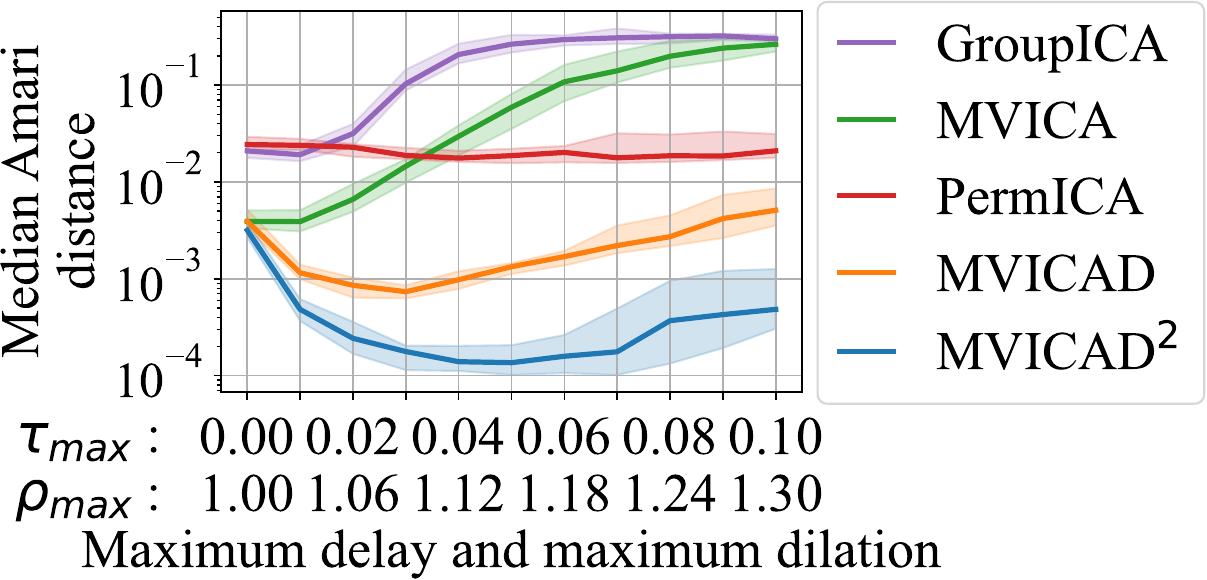}}
\caption{
\centering
Amari distance of multiple methods with respect to the maximum delay $\tau_{\max}$ and the maximum dilation $\rho_{\max}$.
We used $30$ different seeds and plotted the median of all seeds.}
\label{fig:amari_distance_wrt_max_dilation_and_max_shift}
\end{figure}

Figure \ref{fig:amari_distance_wrt_max_dilation_and_max_shift} was obtained by jointly varying the quantity of delays and dilations introduced in the model.
We chose to vary $\tau_{\max}$ from 0 to 0.10 (resp. $\rho_{\max}$ from 1 to 1.30) because, in the context of MEG data and with a sampling rate of 1000~Hz, the maximum delay observed (resp. the maximum dilation observed) is approximately 0.06 (resp. 1.4) for a visual stimulus, and 0.02 (resp. 1.25) for an auditory stimulus \cite{price2017age}.
We observe that MVICA, MVICAD, and MVICAD\textsuperscript{2} approximately start from the same point, but only MVICAD and MVICAD\textsuperscript{2} improve when the quantity of delays and dilations increases.
On the other hand, MVICA and GroupICA suffer from the addition of temporal variations.
As for PermICA, it estimates sources independently for each view, so it is not affected by $\tau_{\max}$ and $\rho_{\max}$, hence the flat curve.
\added{A similar figure where we systematically vary the strength of delays (resp. dilations) while removing dilations (resp. delays) is reported in Appendix~\ref{appendix:expe_without_delay_or_dilation}.}

\begin{figure*}[!t]
\centerline{\includegraphics[width=2.03\columnwidth]{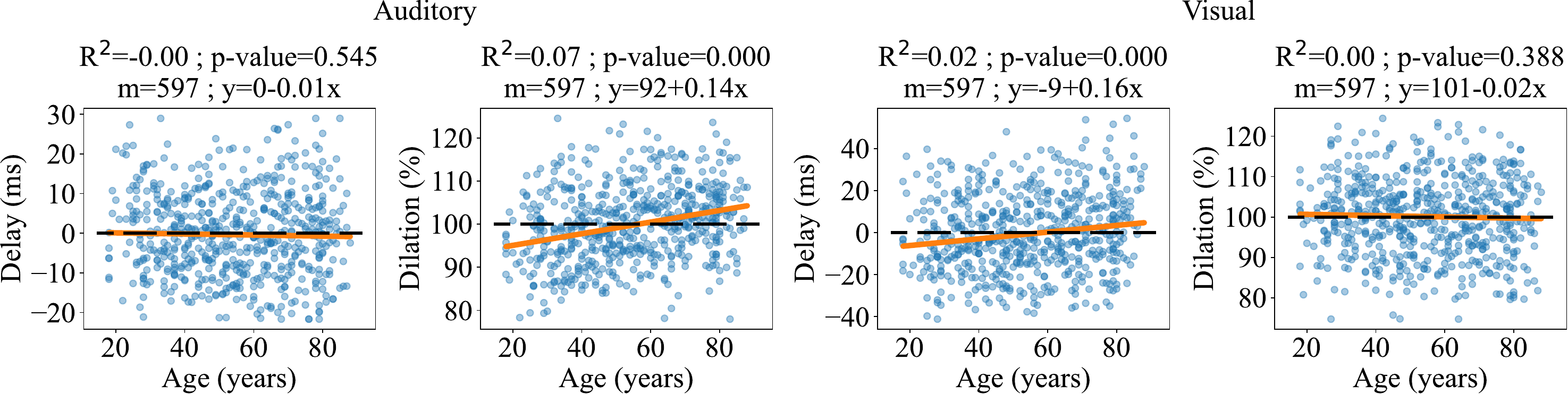}}
\caption{
\centering
Scatter plots of estimated delays and dilations VS ages of the subjects for the auditory task (both figures on the left) and the visual task (both figures on the right).}
\label{fig:scatter_plots_real_data}
\end{figure*}

%% file: content/06_real_data_expes.tex
\section{Real data experiments}\label{sec:real_data_experiments}

\subsection{Cam-CAN dataset}

The Cam-CAN dataset~\cite{taylor2017cambridge} is a large and comprehensive dataset of neuroimaging, cognitive, and demographic data collected from a group of healthy adults.
The dataset includes data from 661 subjects, ranging in age from 18 to 88 years old, with equal numbers of participants in each 10-year age range.
The subjects were recruited from the general population in the Cambridgeshire area of the UK, with the aim of recruiting a sample that was representative of the local population in terms of age, sex, and education level.

The MEG data in the Stage 2 repository includes recordings of brain activity using a whole-head 306 channel Elekta Neuromag Vectorview system.
It contains data from several cognitive tasks, including visual and auditory tasks for which subjects were presented a series of stimuli.

\subsection{Preprocessing}\label{ssec:preprocessing}

We focus on both visual and auditory tasks. Data were preprocessed using a Maxwell filter, a band-pass filter between 0.1 and 40~Hz, and an SSP filter. Trials with gradiometer amplitudes exceeding $3 \times 10^{-10}$~T or magnetometer amplitudes exceeding $1 \times 10^{-11}$~T were excluded. Additionally, six subjects were excluded due to unavailable files (subject IDs: CC120208, CC210023, CC210051, CC520562, CC621118, CC723197), resulting in a final dataset of 597 subjects. To obtain representative time courses, data were averaged across stimuli to retain only $n_{\text{concat}} = 2$ representative time periods. We then extracted epochs of 0.7 seconds, from $t = -0.2$~s to $t = 0.5$~s relative to stimulus onset ($t = 0$). The sampling rate was 1000~Hz, yielding 701 time samples per epoch.

\subsection{Real data results}

One can notice that the metrics $d_{\text{Amari}}$, $d_{\text{delays}}$, and $d_{\text{dilations}}$ depend on true parameters that are usually unknown, so they cannot be used on real data.
Instead, we can verify whether the estimated time delays and dilations are coherent by checking if their correlation with the age of the subjects produces known results in the literature \cite{price2017age}.

We compare our results with \cite{price2017age}.
In this paper, the authors used MEG data of hundreds of subjects that were presented series of visual and auditory stimuli.
They applied PCA on each individual data and only kept the first component.
Then, they estimated how much this first component was delayed and dilated, compared to the average, and \added{examined how these temporal parameters varied with age.}
\added{Dilation was modeled around a stationary point fixed at $t = 50$ ms, as this value corresponds to “the typical latency for information to reach sensory cortices.”} \cite{price2017age}
\added{Note that comparison with existing ICA-based models was not applicable, as they do not estimate subject-specific delays and dilations—except for MVICAD, which models delays only.}

We reproduced this experiment in Figure \ref{fig:scatter_plots_real_data}.
We applied PCA with 10 components to both the visual and auditory datasets described in Section~\ref{ssec:preprocessing}, each containing 597 subjects. This resulted in two datasets of shape $(597,\ 10,\ 2,\ 701)$.
\added{Using averaged epochs rather than raw data greatly reduces the dimensionality required to capture meaningful brain activity, with just 10 components explaining 87\% of the variance.}
Then, we used our method to estimate sources, delays, and dilations\added{, which required approximately 1 hour and 45 minutes on 5 CPU cores}.
Some of the \added{10} estimated sources corresponded to noise and not brain activations, so they \added{were less likely to exhibit meaningful correlations between delays/dilations and age.}
Therefore, to identify representative sources across subjects, we computed, for each source, the average squared difference between individual subject signals and the group-average signal. We retained the three sources with the lowest values—i.e., those most consistent across subjects. Delays and dilations were then averaged over these three sources, and only these aggregated values were retained.

As in \cite{price2017age}, 
\added{we find no significant correlation between age and delays in the auditory task ($p = 0.55$), nor between age and dilations in the visual task ($p = 0.39$). However, we do find highly significant correlation between age and dilations in the auditory task ($p < 10^{-3}$) and between age and delays in the visual task ($p < 10^{-3}$).}
\added{These results align closely with those reported in} \cite{price2017age}.\added{
These observations were consistent across runs with different dimensionalities (8 or 10 components) and stationary points around which dilation occurs ($t = 50$ or $60$ ms), further supporting the method's robustness on real MEG data}

%% file: content/07_discussion.tex
\section{Discussion}

Our method MVICAD$^2$ is designed to learn two types of representations of data: representations that are common across views, and others that are view-specific. 
Results of synthetic experiments show that MVICAD$^2$ outperforms state-of-the-art methods in many scenarios by an order of magnitude (Figures~\ref{fig:synthetic_expes_amari_distance} and~\ref{fig:amari_distance_wrt_max_dilation_and_max_shift}).
Experiments on the Cam-CAN dataset reveal that our method learns delays and dilations for each participant that correlate with their age, suggesting its ability to capture meaningful time variability not modeled by other multi-view ICA algorithms.
However, MVICAD\textsuperscript{2} remains challenged in regimes with high noise levels or a large number of sources to estimate (see Section~\ref{ssec:synthetic_results}); improving robustness in these settings is a direction for future work. \added{Another limitation is that the method is currently tailored to task-evoked responses and may not generalize well to spontaneous activity, which lacks a clear temporal anchor—addressing this would require handling broader forms of temporal variability, such as general time warping. Future work also includes applying the method beyond MEG, particularly to EEG data, and evaluating its utility in a wider range of neuroscientific scenarios.}

%% file: content/08_conclusion.tex
\section*{Conclusion}

We have proposed an unsupervised algorithm, named Multi-View ICA with Delays and Dilations (MVICAD$^2$), that reveals latent sources observed through different views.
This algorithm goes beyond the Multi-View ICA model by allowing latent sources to temporally differ from one view to another.
The temporal variation is modeled as a delay and a dilation, since these effects both appear in neuroimaging group studies.
We proved that our model is identifiable, derived an approximation of its likelihood in closed-form, and proposed regularization and optimization techniques to maximize it efficiently.
We demonstrated, using various synthetic experiments, the superior performance of our method, compared to state-of-the-art multi-view ICA algorithms.
We further validated the usefulness of MVICAD$^2$ for neuroimaging group studies on MEG data, by showing that estimated delays and dilations are correlated to age.
Our method is not specific to neuroimaging data and could be relevant to other observational sciences like genomics or astrophysics where ICA is already widely used.

%% file: content/09_acknowledgment.tex
\section*{Acknowledgment}

Data used in this work were provided by the Cambridge Centre for Ageing and Neuroscience (CamCAN), funded by the UK Biotechnology and Biological Sciences Research Council (BB/H008217/1), the UK Medical Research Council, and the University of Cambridge.

%% file: content/10_appendix.tex
\section{Identifiability proof}
\label{appendix:identifiability_proof}

\begin{proof}
This proof is based on the one of \cite{heurtebise2023multiview}.
Using the assumptions of our identifiability theorem, we have, $\forall i \in [\![m]\!]$, $\forall t \in \bbR$,
\begin{equation}
    \bX^i(t) = A^i (\overrightarrow{\cT}_{\btau^i, \brho^i}(\bS)(t) + \bN^i(t))
\end{equation}
and
\begin{equation}
    \bX^i(t) = A'^i (\overrightarrow{\cT}_{\btau'^i, \brho'^i}(\bS')(t) + \bN'^i(t)) \enspace .
\end{equation}

We assumed that sources were non-Gaussian and independent from one to another and across time.
So, $\overrightarrow{\cT}_{\btau^i, \brho^i}(\bS)$ and $\overrightarrow{\cT}_{\btau'^i, \brho'^i}(\bS')$ have non-Gaussian independent components.
Moreover, $\bN^i$ and $\bN'^i$ have Gaussian independent components.
So, $\overrightarrow{\cT}_{\btau^i, \brho^i}(\bS) + \bN^i$ and $\overrightarrow{\cT}_{\btau'^i, \brho'^i}(\bS') + \bN'^i$ have non-Gaussian independent components.
Following \cite{comon1994independent}, Theorem $11$, there exists a scale-permutation matrix $P^i$ such that $A'^i = A^i P^i$.

As a consequence, and since $A^i$ is invertible, we have, $\forall i \in [\![m]\!]$,
\begin{equation}\label{eq:utils_identifiability}
    \overrightarrow{\cT}_{\btau^i, \brho^i}(\bS) + \bN^i = P^i (\overrightarrow{\cT}_{\btau'^i, \brho'^i}(\bS') + \bN'^i) \enspace .
\end{equation}
We define in \eqref{eq:utils_identifiability} the product between matrix $P^i \in \bbR^{p \times p}$ and multivariate signal $\overrightarrow{\cT}_{\btau'^i, \brho'^i}(\bS') + \bN'^i : \bbR \rightarrow \bbR^p$ by considering that it is done pointwise for each time $t$.
From now on, we allow such products.
By applying $\overleftarrow{\cT}_{\btau^i, \brho^i}$ on both sides of \eqref{eq:utils_identifiability}, we get

\begin{align}
    \bS + \overleftarrow{\cT}_{\btau^i, \brho^i}(\bN^i) 
    & \! = \! \overleftarrow{\cT}_{\btau^i, \brho^i} \left( P^i (\overrightarrow{\cT}_{\btau'^i, \brho'^i}(\bS') + \bN'^i) \right) \nonumber \\
    & \! = \! P^i \overleftarrow{\cT}_{(P^i)^{\top} \btau^i, (P^i)^{\top} \brho^i} (\overrightarrow{\cT}_{\btau'^i, \brho'^i}(\bS') + \bN'^i) \, ,
\end{align}
where the last equality comes from the fact that, for any vectors $\btau, \brho \in \bbR^p$, any multivariate signal $\bS : \bbR \rightarrow \bbR^p$, and any scale-permutation matrix $P \in \bbR^{p \times p}$, we have 
\begin{equation}
    \overleftarrow{\cT}_{\btau, \brho} (P \bS) = P \overleftarrow{\cT}_{P^\top \btau, P^\top \brho} (\bS) \enspace .
\end{equation}
Furthermore, for any vectors $\btau, \brho, \btau', \brho' \in \bbR^p$, any multivariate signal $\bS : \bbR \rightarrow \bbR^p$, and any time $\bt \in \bbR^p$, the composition of $\overleftarrow{\cT}_{\btau, \brho}$ and $\overrightarrow{\cT}_{\btau', \brho'}$ gives
\begin{align}
    \overleftarrow{\cT}_{\btau, \brho} \circ \overrightarrow{\cT}_{\btau', \brho'}(\bS)(\bt) 
    &= \bS \left( \frac{1}{\brho} \odot (\brho' \odot (\bt - \btau')) + \btau \right) \nonumber \\
    &= \bS \left( \frac{\brho'}{\brho} \odot \bt - \frac{\brho' \odot \btau'}{\brho} + \btau \right) \nonumber \\
    &= \overleftarrow{\cT}_{\btau - \frac{\brho' \odot \btau'}{\brho}, \frac{\brho}{\brho'}} (\bS)(\bt) \enspace ,
\end{align}
where $\frac{\brho'}{\brho}$ and $\frac{\brho' \odot \btau'}{\brho}$ use the element-wise division. Consequently,
\begin{align}\label{eq:utils_identifiability_2}
    \bS + \overleftarrow{\cT}_{\btau^i, \brho^i}(\bN^i) 
    &= P^i \overleftarrow{\cT}_{(P^i)^{\top} \btau^i - \frac{\brho'^i \odot \btau'^i}{(P^i)^{\top} \brho^i}, \frac{(P^i)^{\top} \brho^i}{\brho'^i}}(\bS') \nonumber \\
    & + P^i \overleftarrow{\cT}_{(P^i)^{\top} \btau^i, (P^i)^{\top} \brho^i} (\bN'^i) \enspace .
\end{align}

From now on, we define for simplicity, $\forall i \in [\![m]\!]$,
\begin{equation}\label{eq:utils_identifiability_definition_3}
    \tilde{\btau}^i := (P^i)^{\top} \btau^i - \frac{\brho'^i \odot \btau'^i}{(P^i)^{\top} \brho^i} \quad \text{and} \quad \tilde{\brho}^i := \frac{(P^i)^{\top} \brho^i}{\brho'^i} \enspace .
\end{equation}
In addition, we define the noises
\begin{equation}\label{eq:utils_definition_1}
    \overleftarrow{\bN}^i := \overleftarrow{\cT}_{\btau^i, \brho^i}(\bN^i) \quad \text{and} \quad \overleftarrow{\bN}'^i_P := P^i \overleftarrow{\cT}_{(P^i)^{\top} \btau^i, (P^i)^{\top} \brho^i} (\bN'^i)
\end{equation}
and we also define
\begin{equation}\label{eq:utils_definition_2}
    \overleftarrow{\bS}'^i := \overleftarrow{\cT}_{\tilde{\btau}^i, \tilde{\brho}^i}(\bS') \enspace .
\end{equation}
Using definitions \eqref{eq:utils_definition_1} and \eqref{eq:utils_definition_2}, we can reformulate \eqref{eq:utils_identifiability_2} as, $\forall i \in [\![m]\!]$,
\begin{equation}
    \bS + \overleftarrow{\bN}^i = P^i \overleftarrow{\bS}'^i + \overleftarrow{\bN}'^i_P \enspace .
\end{equation}

We focus on subject $1$ and subject $i \neq 1$. We have
\begin{equation}
    \bS + \overleftarrow{\bN}^1 - \left( \bS + \overleftarrow{\bN}^i \right) = P^1 \overleftarrow{\bS}'^1 + \overleftarrow{\bN}'^1_P - \left( P^i \overleftarrow{\bS}'^i + \overleftarrow{\bN}'^i_P \right) \enspace .
\end{equation}
Thus,
\begin{equation}\label{eq:right_gauss}
    P^1 \overleftarrow{\bS}'^1 - P^i \overleftarrow{\bS}'^i = \overleftarrow{\bN}^1 - \overleftarrow{\bN}^i + \overleftarrow{\bN}'^i_P - \overleftarrow{\bN}'^1_P \enspace .
\end{equation}
The right-hand side of \eqref{eq:right_gauss} is a linear combination of Gaussian random variables, which implies that $P^1 \overleftarrow{\bS}'^1 - P^i \overleftarrow{\bS}'^i$ is also Gaussian.

Let us show that $\tilde{\btau}^1 = \tilde{\btau}^i$ and $\tilde{\brho}^1 = \tilde{\brho}^i$.
By contradiction, suppose that $\tilde{\btau}^1 \neq \tilde{\btau}^i$ or $\tilde{\brho}^1 \neq \tilde{\brho}^i$, i.e. $\exists j \in [\![p]\!]$, $\tilde{\tau}^1_j \neq \tilde{\tau}^i_j$ or $\tilde{\rho}^1_j \neq \tilde{\rho}^i_j$.
Then, for all $t \in \bbR$ (except maybe for one time point), we have 
\begin{equation}\label{eq:utils_identifiability_3}
    \frac{1}{\tilde{\rho}^1_j} t + \tilde{\tau}^1_j \neq \frac{1}{\tilde{\rho}^i_j} t + \tilde{\tau}^i_j \enspace .
\end{equation}
We fix such a time $t$. Let us define for simplicity
\begin{equation}
    \bu := \overleftarrow{\bS}'^1 (t) \in \bbR^p \quad \text{and} \quad \bv := \overleftarrow{\bS}'^i (t) \in \bbR^p \enspace .
\end{equation}
From \eqref{eq:utils_definition_2}, \eqref{eq:utils_identifiability_3}, and the fact that $\bS'$ is independent across time, we deduce that $u_j$ and $v_j$ are independent, where $u_j$ (resp. $v_j$) is the $j$-th component of $\bu$ (resp. $\bv$).
Since the components of $\bv$ are independent from one to another and $u_j$ is independent from $v_j$, then $u_j$ is independent from all the components of $\bv$.

We know that $P^1$ is a scale-permutation matrix, so it has exactly one non-zero value in each of its columns.
Let us call $j'$ the row number of the non-zero value of its $j$-th column.
Matrix $P^1$ moves $u_j$ from the $j$-th row to the $j'$-th row, as follows: $(P^1 \bu)_{j'} = P^1_{j' j} u_j$.
Since $u_j$ is independent of all the entries of $\bv$, we have, in particular, that $(P^1 \bu)_{j'}$ is independent of $(P^i \bv)_{j'}$.
In other words, we proved that $(P^1 \overleftarrow{\bS}'^1 (t))_{j'}$ is independent from $(P^i \overleftarrow{\bS}'^i (t))_{j'}$.

Since the whole multivariate signal $P^1 \overleftarrow{\bS}'^1 - P^i \overleftarrow{\bS}'^i$ is Gaussian, we deduce that, in particular, $(P^1 \overleftarrow{\bS}'^1 (t))_{j'} - (P^i \overleftarrow{\bS}'^i (t))_{j'}$ is Gaussian too.
So, by Lévy-Cramér's theorem, we should have that $\overleftarrow{\bS}'^1_{j'} (t)$ and $\overleftarrow{\bS}'^i_{j'} (t)$ are Gaussian.
But this is absurd, given that all the components of $\bS'$ are non-Gaussian.
So, $\tilde{\btau}^1 = \tilde{\btau}^i$ and $\tilde{\brho}^1 = \tilde{\brho}^i$.

In other words, there exists vectors $\btau_\star$ and $\brho_\star$ such that, $\forall i \in [\![m]\!]$,
\begin{equation}
    \tilde{\btau}^1 = \tilde{\btau}^i =: \btau_\star \quad \text{and} \quad \tilde{\brho}^1 = \tilde{\brho}^i =: \brho_\star \enspace .
\end{equation}
Using the definitions of $\tilde{\btau}^i$ and $\tilde{\brho}^i$ in \eqref{eq:utils_identifiability_definition_3}, it follows that, $\forall i \in [\![m]\!]$,
\begin{equation}
    \btau'^i = ((P^i)^{\top} \btau^i - \btau_\star) \odot \brho_\star \quad \text{and} \quad \brho'^i = \frac{(P^i)^{\top} \brho^i}{\brho_\star} \enspace .
\end{equation}

We recall that, $\forall i \in [\![m]\!]$, $\overleftarrow{\bS}'^i = \overleftarrow{\cT}_{\tilde{\btau}^i, \tilde{\brho}^i}(\bS')$.
Given that $\tilde{\btau}^1 = \tilde{\btau}^i$ and $\tilde{\brho}^1 = \tilde{\brho}^i$, we define
\begin{equation}
    \tilde{\bS} := \overleftarrow{\bS}'^1 = \overleftarrow{\bS}'^i \enspace .
\end{equation}
We know that $P^1 \overleftarrow{\bS}'^1 - P^i \overleftarrow{\bS}'^i$ is Gaussian.
In other words, $(P^1 - P^i) \tilde{\bS}$ is Gaussian.
This only holds if $P^1 = P^i$.
Therefore, the matrices $P^i$ are all equal, and there exists a scale-permutation matrix $P_\star \in \bbR^{p \times p}$ such that, $\forall i \in [\![m]\!]$, $A'^i = A^i P_\star$.

In conclusion, we proved that there exists a scale-permutation matrix $P_\star$ and vectors $\btau_\star$ and $\brho_\star$ such that, $\forall i \in [\![m]\!]$,
\begin{equation}\label{app:eq:identifiability_conclusion}
    A'^i = A^i P_\star \; , \ \btau'^i = (P_\star^{\top} \btau^i - \btau_\star) \odot \brho_\star \ \text{and} \ \brho'^i = \frac{P_\star^{\top} \brho^i}{\brho_\star} \enspace .
\end{equation}

\end{proof}

\newpage

\section{Deriving the approximate negative log-likelihood}
\label{appendix:NLL_computations}

\begin{proof}
Let us derive the approximation of the negative log-likelihood (NLL) of our model.
Recall that our model is, $\forall i \in [\![m]\!]$ and $\forall t \in \bbR$,
\begin{equation}\label{eq:utils_recall_model}
    \bX^i(t) = A^i \left( \overrightarrow{\cT}_{\btau^i, \brho^i}(\bS)(t) + \bN^i(t) \right) \enspace .
\end{equation}
Let us define
\begin{equation}\label{eq:definition_Zt}
    \bZ^i = \overrightarrow{\cT}_{\btau^i, \brho^i}(\bS) \enspace .
\end{equation}
By change of variables, 
we have, from \eqref{eq:utils_recall_model} and \eqref{eq:definition_Zt}, that
\begin{equation}
    \bN^i(t) = W^i \bX^i(t) - \bZ^i(t) \enspace ,
\end{equation}
where $W^i = (A^i)^{-1}$. So, the distribution of $\bX^i(t)$ conditioned on $\bZ^i(t)$, is
\begin{align}
    \bbP_{\bX^i} (\bX^i(t) | \bZ^i(t); W^i, \btau^i, \brho^i) 
    =
    |\det(W^i)| \bbP_{\bN^i} (W^i \bX^i(t) - \bZ^i(t)) \enspace ,
\end{align}
where $\bbP_{\bN^i}$ is the distribution of $\bN^i(t)$, i.e. $\cN(0_p, \sigma^2 I_p)$.
Thus, 
\begin{align}
    \bbP_{\bN^i} (W^i \bX^i(t) - \bZ^i(t))
    =
    (2 \pi \sigma^2)^{-\frac{p}{2}} \exp \left( - \frac{\left\| W^i \bX^i(t) - \bZ^i(t) \right\|^2}{2 \sigma^2} \right) \enspace .
\end{align}

We define $\cX = (\bX^1, \dots, \bX^m)$ and $\cZ = (\bZ^1, \dots, \bZ^m)$.
Let use use the notations $\cX(t) = (\bX^1(t), \dots, \bX^m(t))$ and $\cZ(t) = (\bZ^1(t), \dots, \bZ^m(t))$, and recall that $\cW = (W^1, \dots, W^m)$, $\btau = (\btau^1, \dots, \btau^m)$, and $\brho = (\brho^1, \dots, \brho^m)$.
We can remark that, conditionally on $\bZ^i(t)$, $W^i$, $\btau^i$, and $\brho^i$, $i=1, \dots, m$, the r.v. $\bX^1(t), \dots, \bX^m(t)$ are mutually independent.
Consequently, we have
\begin{align}
    \bbP_{\cX}(\cX(t) | \cZ(t); \cW, \btau, \brho)
    =
    \prod_{i=1}^m |\det(W^i)| (2 \pi \sigma^2)^{-\frac{p}{2}} \exp \left( - \frac{\left\| W^i \bX^i(t) - \bZ^i(t) \right\|^2}{2 \sigma^2} \right) \, .
\end{align}
To derive the joint distribution of $\cX(t)$ and $\cZ(t)$, conditioned on $\cW$, $\btau$, and $\brho$, we multiply by the distribution of $\cZ(t)$, as follows:
\begin{align}\label{eq:joint_distribution}
    \bbP_{\cX}(\cX(t), \cZ(t) | \cW, \btau, \brho) 
    =
    \bbP_{\cZ}(\cZ(t)) \! \prod_{i=1}^m \! |\det(W^i)| (2 \pi \sigma^2)^{-\frac{p}{2}}
    \exp \left( - \frac{\left\| W^i \bX^i(t) - \bZ^i(t) \right\|^2}{2 \sigma^2} \right) \enspace .
\end{align}
Note that the distribution of $\cZ(t)$ is the joint distribution of $(\bZ^1(t), \dots, \bZ^m(t))$, so it can be written $\bbP_{(\bZ^1, \dots, \bZ^m)}(\bZ^1(t), \dots, \bZ^m(t))$.
By integrating \eqref{eq:joint_distribution} over $\cZ(t)$, we derive the likelihood at time $t$ of our model:
\begin{align}
    \bbP_{\cX} (\cX(t) | \cW, \btau, \brho)
    & = 
    \int_{\bz^1_t, \dots, \bz^m_t \in \bbR^p} \bbP_{(\bZ^1, \dots, \bZ^m)}(\bz^1_t, \dots, \bz^m_t) \prod_{i=1}^m |\det(W^i)| \nonumber \\
    & \times
    (2 \pi \sigma^2)^{-\frac{p}{2}} \exp \left( - \frac{\left\| W^i \bX^i(t) - \bz^i_t \right\|^2}{2 \sigma^2} \right) d\bz^1_t \dots d\bz^m_t \enspace .
\end{align}

Thus, the NLL at time $t$ is
\begin{align}
    \cL_t(\cW, \btau, \brho) &= -\sum_{i=1}^m \log|\det(W^i)| + \frac{mp}{2}\log(2\pi \sigma^2) \nonumber \\
    & -\log \left( \int_{\bz^1_t, \dots, \bz^m_t \in \bbR^p} \bbP_{(\bZ^1, \dots, \bZ^m)}(\bz^1_t, \dots, \bz^m_t) \right.
    \left. \! \prod_{i=1}^m \! \exp \! \left( \! - \frac{\left\| W^i \bX^i(t) - \bz^i_t \right\|^2}{2 \sigma^2} \! \right) d\bz^1_t \dots d\bz^m_t \right) ,
\end{align}
where the integrand factorizes in the previous integral.
Indeed, we know that, $\forall t \in \bbR$, the components of $\bS(t)$ are independent of each other and the $\bS(t)$ are independent across time $t$. 
So, the $(\bZ^1_j, \dots, \bZ^m_j)$, for $j=1, \dots, p$, are independent of each other, where $\bZ^i_j$ is the $j$-th component of $\bZ^i$.
Thus, the integrand can be written
\begin{align}
    \bbP_{(\bZ^1, \dots, \bZ^m)}(\bz^1_t,& \dots, \bz^m_t) \prod_{i=1}^m \exp \left( - \frac{\left\| W^i \bX^i(t) - \bz^i_t \right\|^2}{2 \sigma^2} \right) \nonumber \\
    & = \prod_{j=1}^p \bbP_{(\bZ^1_j, \dots, \bZ^m_j)}(z^1_{j, t}, \dots, z^m_{j, t}) \prod_{i=1}^m \exp \left( - \frac{\left( (\bw^i_j)^\top \bX^i(t) - z^i_{j, t} \right)^2}{2 \sigma^2} \right) \enspace , \label{eq:integrand}
\end{align}
where $z^i_{j, t}$ is a realization of the $j$-th component of $\bZ^i(t)$, and $\bw^i_j$ is the $j$-th row of $W^i$.

Assume that, $\forall j \in [\![p]\!]$, $\forall i_1, i_2 \in [\![m]\!]$, $i_1 \neq i_2$, we have $\tau^{i_1}_j \neq \tau^{i_2}_j$ or $\rho^{i_1}_j \neq \rho^{i_2}_j$, which is almost sure if $\btau$ and $\brho$ are drawn uniformly.
Then, for all $t \in \bbR$ (except maybe for one value of $t$ where the two affine lines cross), we have
\begin{equation}\label{eq:utils_NLL}
    \rho^{i_1}_j (t - \tau^{i_1}_j) \neq \rho^{i_2}_j (t - \tau^{i_2}_j) \enspace .
\end{equation}
Given \eqref{eq:definition_Zt}, \eqref{eq:utils_NLL}, and the fact that the $\bS_j(t)$ are i.i.d. across time, we get that, for all $t$ (except maybe for one of them), $\bZ^{i_1}_j(t)$ and $\bZ^{i_2}_j(t)$ are independent.
So, for almost all $t \in \bbR$, we have
\begin{equation}
    \bbP_{(\bZ^1_j, \dots, \bZ^m_j)}(z^1_{j, t}, \dots, z^m_{j, t}) = \prod_{i=1}^m \bbP_{\bZ_j}(z^i_{j, t}) \enspace .
\end{equation}
As a consequence, for almost all $t \in \bbR$,
\begin{equation}
    \eqref{eq:integrand} = \prod_{i=1}^m \prod_{j=1}^p \bbP_{\bZ_j}(z^i_{j, t}) \exp \left( - \frac{\left( (\bw^i_j)^\top \bX^i(t) - z^i_{j, t} \right)^2}{2 \sigma^2} \right) \enspace .
\end{equation}

It follows that, up to the constant $\frac{mp}{2}\log(2\pi \sigma^2)$, and for almost all $t \in \bbR$, the NLL at time $t$ is
\begin{align}
    \cL_t(\cW, \btau, \brho) = -\sum_{i=1}^m \log|\det(W^i)| 
    -\sum_{i=1}^m \sum_{j=1}^p \log \left( \int_{z^i_{j, t} \in \bbR} \bbP_{\bZ_j}(z^i_{j, t}) \right.
    \left. \exp \left( - \frac{\left( (\bw^i_j)^\top \bX^i(t) - z^i_{j, t} \right)^2}{2 \sigma^2} \right) dz^i_{j, t} \right) \enspace .
\end{align}

Let $T > 0$.
The NLL averaged over time is
\begin{align}
    \cL(\cW, \btau, \brho) &= \int_{t=0}^T \cL_t(\cW, \btau, \brho) dt \nonumber \\
    & = -T \sum_{i=1}^m \log|\det(W^i)| -\sum_{i=1}^m \sum_{j=1}^p \delta^i_j \enspace , \label{eq:L_over_time} 
\end{align}
where $\delta^i_j \in \bbR$ is defined as
\begin{align}
    \delta^i_j = \int_{t=0}^T \log \left( \int_{z^i_{j, t} \in \bbR} \bbP_{\bZ_j}(z^i_{j, t}) \right.
    \left. \exp \left( - \frac{\left( (\bw^i_j)^\top \bX^i(t) - z^i_{j, t} \right)^2}{2 \sigma^2} \right) dz^i_{j, t} \right) dt \enspace .
\end{align}

We want to use the change of variable that corresponds to the $j$-th component of the operator $\overleftarrow{\cT}_{\btau^i, \brho^i}$ defined as, $\forall \btau^i, \brho^i \in \bbR^p$ and $\forall \bt \in \bbR^p$,
\begin{equation}
    \overleftarrow{\cT}_{\btau^i, \brho^i}(\bS)(\bt) = \bS \left( \frac{1}{\brho^i} \odot \bt + \btau^i \right) \enspace .
\end{equation}
Let $\tilde{\bt}^i = \brho^i \odot (t \mathbb{1}_p - \btau^i)$, which implies that $t \mathbb{1}_p = \frac{1}{\brho^i} \odot \tilde{\bt}^i + \btau^i$.
By the definition of $\bZ^i$ given in \eqref{eq:definition_Zt}, we have, $\forall t \in \bbR$,
\begin{align}\label{eq:T_left}
    \bZ^i(t) = \bZ^i(t \mathbb{1}_p) &= \bZ^i \left( \frac{1}{\brho^i} \odot \tilde{\bt}^i + \btau^i \right) \nonumber \\
    &= \overleftarrow{\cT}_{\btau^i, \brho^i}(\bZ^i)(\tilde{\bt}^i) = \bS(\tilde{\bt}^i) \enspace .
\end{align}
Let $\tilde{t} = \rho^i_j (t - \tau^i_j)$, where we will drop the dependence on $i$ and $j$ for simplicity.
In addition, $\forall i \in [\![m]\!]$, let us define $\bS^i = (\bS^i(t))_{t \in \bbR}$, where $\bS^i(t) := W^i \bX^i(t) \in \bbR^p$.
The $j$-th component of $\bS^i$ is $\bS^i_j = (\bw^i_j)^\top \bX^i$, so that it implicitly depends on $\bw^i_j$.
Note that $S^i$ are the estimated sources for view $i$, and are different from the true shared sources $S$.
Also note that, in this definition, we can multiply $\bX^i$ by $W^i$ or $(\bw^i_j)^\top$ either before or after fixing the time $t$.
Using the same idea as in \eqref{eq:T_left} but in one dimension, the change of variable gives
\begin{align}
    \delta^i_j = \int_{\tilde{t}=-\rho^i_j \tau^i_j}^{\rho^i_j (T - \tau^i_j)} \log \left( \int_{s_{j, \tilde{t}} \in \bbR} \bbP_{\bS_j}(s_{j, \tilde{t}}) \right.
    \left. \exp \left( - \frac{\left( \bS^i_j \left( \frac{1}{\rho^i_j} \tilde{t} + \tau^i_j \right) - s_{j, \tilde{t}} \right)^2}{2 \sigma^2} \right) ds_{j, \tilde{t}} \right) \frac{1}{\rho^i_j} d\tilde{t} \enspace .
\end{align}

Let us define the following approximation of $\delta^i_j$:
\begin{align}\label{eq:B^i_j_tilde}
    \tilde{\delta}^i_j = \int_{\tilde{t}=0}^T \log \left( \int_{s_{j, \tilde{t}} \in \bbR} \bbP_{\bS_j}(s_{j, \tilde{t}}) \right.
    \left. \exp \left( - \frac{\left( \bS^i_j \left( \frac{1}{\rho^i_j} \tilde{t} + \tau^i_j \right) - s_{j, \tilde{t}} \right)^2}{2 \sigma^2} \right) ds_{j, \tilde{t}} \right) d\tilde{t}
\end{align}
where we modified the range of the integral and removed the multiplicative factor of $1 / \rho^i_j$ in the integrand.
From \eqref{eq:L_over_time} and \eqref{eq:B^i_j_tilde}, let us also define an approximation of the NLL integrated over time:
\begin{equation}
    \tilde{\cL}(\cW, \btau, \brho) = -T \sum_{i=1}^m \log|\det(W^i)| -\sum_{i=1}^m \sum_{j=1}^p \tilde{\delta}^i_j \enspace .
\end{equation}
Since $\forall i \in [\![m]\!], \forall j \in [\![p]\!]$, $\tilde{\delta}^i_j \rightarrow \delta^i_j$ as soon as $\tau^i_j \rightarrow 0$ and $\rho^i_j \rightarrow 1$, we infer that $\tilde{\cL}(\cW, \btau, \brho) \rightarrow \cL(\cW, \btau, \brho)$ as soon as $\btau \rightarrow \mathbf{0}$ and $\brho \rightarrow \mathbf{1}$.

In the following, we replace $\tilde{t}$ with $t$ for simplicity.
One can notice that, in \eqref{eq:B^i_j_tilde}, $\bS^i_j \left( \frac{1}{\rho^i_j} t + \tau^i_j \right)$ corresponds to the $j$-th component of $\bY^i(t)$, which is defined in the paper as $\bY^i(t) = \overleftarrow{\cT}_{\btau^i, \brho^i}(W^i \bX^i)(t) \in \bbR^p$.
Let $y^i_{j, t} := \bY^i_j(t) \in \bbR$.
The approximate-NLL is
\begin{align}
    \tilde{\cL}(\cW, \btau, \brho)
    =
    -T \sum_{i=1}^m \log|\det(W^i)|
    - \int_{t=0}^T \log \left( \prod_{j=1}^p \int_{s_{j, t} \in \bbR} \bbP_{\bS_j}(s_{j, t}) \right.
    \left. \exp \left( - \frac{\sum_{i=1}^m \left( y^i_{j, t} - s_{j, t} \right)^2}{2 \sigma^2} \right) ds_{j, t} \right) dt \label{eq:NLL_approx} \enspace .
\end{align}

Let us focus on the integral over $s_{j, t}$, then fix $j$ and $t$, and drop them for simplicity.
We need to solve
\begin{equation}\label{eq:int_drop_j}
    \int_s \bbP_{\bS}(s) \exp \left( - \frac{\sum_{i=1}^m (y^i - s)^2}{2 \sigma^2} \right) ds \enspace .
\end{equation}
Following~\cite{richard2020modeling}, we wish to rewrite \eqref{eq:NLL_approx} in terms of the sources averaged across views.
To do so, we use the following decomposition.
For any $y \in \bbR$,
\begin{align}
    \sum_{i=1}^m (y^i - s)^2
    &= \sum_{i=1}^m (y^i - y + y - s)^2 \nonumber \\
    &= m (y - s)^2 + \sum_{i=1}^m (y^i - y)^2
    + 2 (y - s) \sum_{i=1}^m (y^i - y) \enspace ,
\end{align}
and $\sum_{i=1}^m (y^i - y)$ vanishes if we take $y = \frac{1}{m} \sum_{i=1}^m y^i =: \overline{y}$.
So,
\begin{align}
    \eqref{eq:int_drop_j}
    &= \int_s \bbP_{\bS}(s) \exp \left( - \frac{m(\overline{y} - s)^2 + \sum_{i=1}^m (y^i - \overline{y})^2}{2 \sigma^2}  \right) ds \nonumber \\
    &= \exp \left( - \frac{\sum_{i=1}^m (y^i - \overline{y})^2}{2 \sigma^2} \right) \int_s \bbP_{\bS}(s) \exp \left( -\frac{m}{2 \sigma^2}(\overline{y} - s)^2 \right) ds \nonumber \\
    &= \exp \left( -\frac{\sum_{i=1}^m (y^i - \overline{y})^2}{2 \sigma^2} \right) \exp(-f(\overline{y})) \enspace , \label{eq:int_simplified_with_f}
\end{align}
by using a change of variable and where $f$ is defined in the paper as
\begin{equation}
    f(\overline{y}) := -\log \left( \int_s \bbP_{\bS}(\overline{y} - s) \exp \left( -\frac{m}{2 \sigma^2} s^2 \right) ds \right) \enspace .
\end{equation}

Putting \eqref{eq:int_simplified_with_f} into \eqref{eq:NLL_approx} gives
\begin{align}
    \tilde{\cL}(\cW, \btau, \brho)
    &=
    -T \sum_{i=1}^m \log|\det(W^i)| - \int_{t=0}^T \log \Bigg( \prod_{j=1}^p \exp (-f(\overline{y}_{j, t})) \exp \left( - \frac{\sum_{i=1}^m (y^i_{j, t} - \overline{y}_{j, t})^2}{2 \sigma^2} \right) \Bigg) dt \nonumber \\
    &= -T \sum_{i=1}^m \log|\det(W^i)| - \int_{t=0}^T \log \Bigg( \exp \Bigg( -\sum_{j=1}^p f(\overline{y}_{j, t}) \Bigg) \exp \left( - \frac{\sum_{i=1}^m \left\| \bY^i(t)  - \overline{\bY}(t) \right\|^2}{2 \sigma^2} \right) \Bigg) dt \enspace , 
\end{align}
where $\overline{\bY}(t)$ is defined in the paper as $\overline{\bY}(t) = \frac{1}{m} \sum_{i=1}^m \bY^i(t)$.
It follows that
\begin{align}
    \tilde{\cL}(\cW, \btau, \brho)
    =
    -T \sum_{i=1}^m \log|\det(W^i)| 
    +
    \int_{t=0}^T f(\overline{\bY}(t)) dt
    +
    \int_{t=0}^T \frac{1}{2 \sigma^2} \sum_{i=1}^m \left\| \bY^i(t)  - \overline{\bY}(t) \right\|^2 dt \enspace .
\end{align}

By dividing by $T$, we get that the averaged approximation of the negative log-likelihood over $[0, T]$ is, up to a factor,
\begin{align}
    \tilde{\cL}(\cW, \btau, \brho)
    =
    -\sum_{i=1}^m \log|\det(W^i)|
    +
    \frac{1}{T} \int_{t=0}^T f \left( \overline{\bY}(t) \right) dt
    +
    \frac{1}{T} \int_{t=0}^T \frac{1}{2 \sigma^2} \sum_{i=1}^m \left\| \bY^i(t) - \overline{\bY}(t) \right\|^2 dt \enspace .
\end{align}
In the paper, we call this approximate-NLL $\cL$ instead of $\tilde{\cL}$.
\end{proof}

\newpage

\section{Details on the regularization term \texorpdfstring{$\cR_1$}{R1} and its scale}
\label{appendix:regularization_R1}

\added{The explicit form of the regularization term $\cR_1$ is}
\begin{align}
    \cR_1 (\btau, \brho) 
    = 
    \sum_{j=1}^p \left(\frac{\overline{\tau}_j}{\tau_{\max}}\right)^2
    + 
    \sum_{j=1}^p \left(\frac{\overline{\rho}_j-1}{\rho_{\max} \ind_{\{\overline{\rho}_j \geq 1\}} + \frac{1}{\rho_{\max}} \ind_{\{\overline{\rho}_j < 1\}} - 1}\right)^2 \enspace ,
\end{align}
\added{where $\overline{\tau}_j = \frac1m \sum_{i=1}^m \tau^i_j$ and $\overline{\rho}_j = \frac1m \sum_{i=1}^m \rho^i_j$ are the average among subjects and for the $j$-th source.
Since for all $j$, $\overline{\tau}_j \in [-\tau_{\max}, \tau_{\max}]$ and $\overline{\rho}_j \in [1/\rho_{\max}, \rho_{\max}]$, we deduce that $\cR_1 (\btau, \brho) \in [0, 2p]$.}

\begin{figure*}[h]
    \centering
    \hspace{-2cm}
    \includegraphics[width=0.3\textwidth]{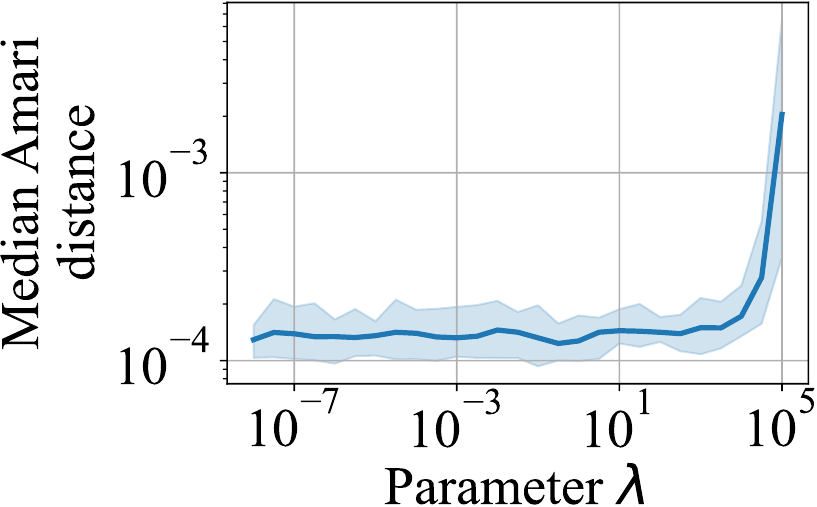}
    \caption{
    \centering 
    Amari distance of our method as a function of the penalization scale $\lambda \geq 0$. \added{The results indicate that performance is largely insensitive to the choice of $\lambda$, supporting the choice of $\lambda=1$ as a default value without the need for tuning.}}
    \label{fig:amari_distance_wrt_penalization_scale}
\end{figure*}

The rest of this section addresses the choice of the hyperparameter $\lambda \geq 0$, which controls the weight given to $\cR_1$ in the objective function. In this experiment, we used a simple setting with $5$ views, $3$ sources, $600$ samples, a maximum delay $\tau_{\max}=0.05$, and a maximum dilation $\rho_{\max}=1.15$. As shown in Figure \ref{fig:amari_distance_wrt_penalization_scale}, the Amari distance is relatively unaffected by $\lambda$ as long as it remains below $10^4$. The sharp increase in error beyond this point can be attributed to $\lambda \cR_1(\btau, \brho)$ dominating the other terms in the loss function, which causes the algorithm to focus primarily on optimizing $\cR_1(\btau, \brho)$. In practice, we set $\lambda = 1$ by default.

\section{Choosing the filter length \texorpdfstring{$l$}{l}}
\label{appendix:filter_length_l}

\added{This section provides guidance on selecting the smoothing filter length $l$ (expressed in number of samples) used in the regularization term $\cR_2(\cY, l)$, defined in~\eqref{eq:regularization_term_2}. To assess its impact on performance, we conducted a synthetic experiment using the same data generation process as in Figure~\ref{fig:amari_distance_wrt_penalization_scale}, and varying $l$ from $1$ to $15$. We also included the case where the $\cR_2(\cY, l)$ is omitted from the loss, shown as the leftmost point in the figure.}

\begin{figure}[h]
    \centering
    \hspace{0.5cm}
    \includegraphics[width=0.5\textwidth]{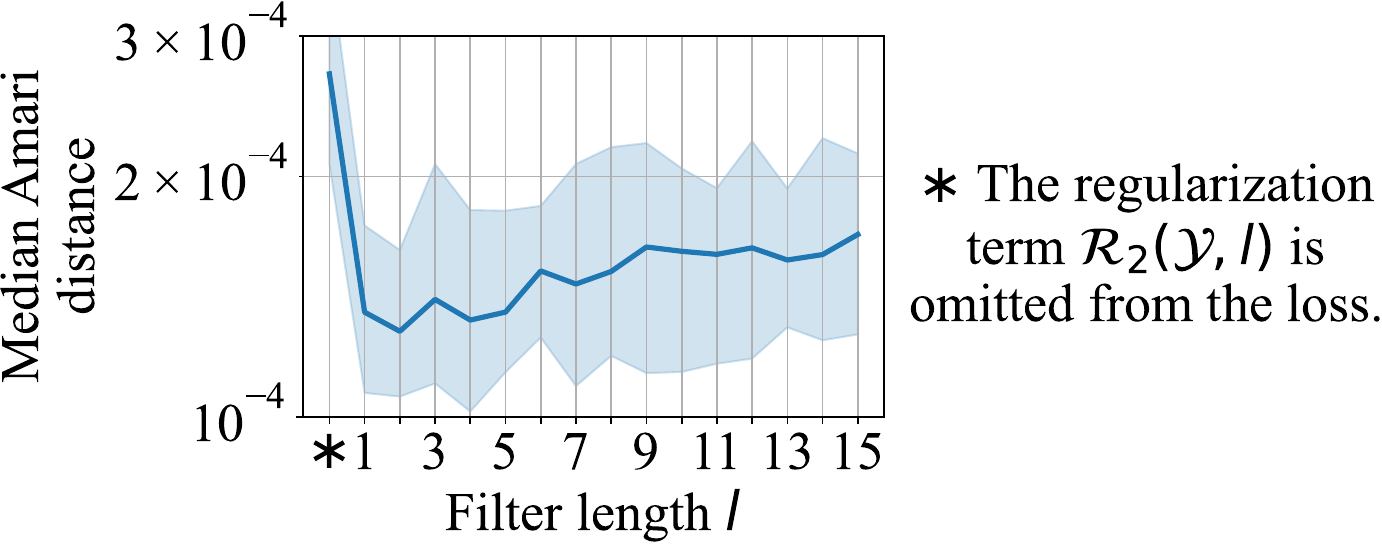}
    \caption{
    \centering 
    \added{Amari distance of our method as a function of the smoothing filter length $l \geq 1$. The results show that the inclusion in the loss of the regularization term $\cR_2$ significantly improves performance. Moreover, the choice of $l$ has little effect—as long as it is not excessively large—justifying the default setting of $l=3$.}}
    \label{fig:amari_distance_wrt_filter_length}
\end{figure}

\added{The results---shown in Figure~\ref{fig:amari_distance_wrt_filter_length}---indicate that, once the term $\cR_2(\cY, l)$ is included, the precise choice of $l$ has little impact on performance, provided it is not too large. Intuitively, this is expected: the role of the regularizer is to guide the optimization of delays and dilations when the signals exhibit high-frequency components, and even minimal smoothing—such as using $l=1$ (i.e., only taking absolute values)—is sufficient. On the other hand, using an excessively wide filter would overly smooth the sources, potentially discarding relevant temporal details and leading to less accurate estimation of the unmixing matrices. Based on this, we set $l=3$ by default.}

\section{Deriving dilations' scale \texorpdfstring{$\Lambda^1$}{Lambda1}}
\label{appendix:parameters_scales_1}

\begin{proof}
Recall that $\overline{\bY} : \bbR \rightarrow \bbR^p$ are the averaged estimated sources and let $\tilde{\btau} = (\tilde{\tau}_1, \ldots, \tilde{\tau}_p)^\top \in \bbR^p$ be delays close to $\mathbf{0}$ and $\tilde{\brho} = (\tilde{\rho}_1, \ldots, \tilde{\rho}_p)^\top \in \bbR^p$ be dilations close to $\mathbf{1}$.
Our goal is to find a vector $\Lambda^1 = (\Lambda_1, \ldots, \Lambda_p)^\top \in \bbR^p$ such that, for each source number $j \in [\![ p ]\!]$, delaying the $j$-th component $\overline{\bY}_j$ by $\Lambda_j \tilde{\tau}_j \in \bbR$ and dilating it by $\tilde{\rho}_j \in \bbR$ have comparable effects on the quantity
\begin{equation}\label{app:eq:loss_lambda}
    \int_0^T  \left\| \overrightarrow{\cT}_{\Lambda^1 \odot \tilde{\btau}, \tilde{\brho}} (\overline{\bY})(t) - \overline{\bY}(t) \right\|^2 dt \enspace .
\end{equation}
To do so, we use the second-order information of \eqref{app:eq:loss_lambda}. More specifically, we want to calculate the Hessian matrix of \eqref{app:eq:loss_lambda} with respect to $\tilde{\btau}$ and $\tilde{\brho}$.

We can rewrite \eqref{app:eq:loss_lambda} as
\begin{equation}
    \eqref{app:eq:loss_lambda}
    =
    \sum_{j=1}^p \int_{0}^T \left( \overrightarrow{\cT}_{\Lambda_j \tilde{\tau}_j, \tilde{\rho}_j} (\overline{\bY}_j)(t) - \overline{\bY}_j(t) \right)^2 dt \enspace ,
\end{equation}
where we observe that the pairs $(\tilde{\tau}_j, \tilde{\rho}_j)_{j \in [\![ p ]\!]}$ have no effect on each other.
Thus, the $2p \times 2p$ Hessian matrix is a block-diagonal matrix, with $p$ blocks of size $2 \times 2$, and we can treat each block separately.

Let us consider the $j$-th block of the Hessian, for a fixed $j \in [\![ p ]\!]$.
For simplicity, we define $\Lambda := \Lambda_j$, $\tau := \tilde{\tau}_j$, $\rho := \tilde{\rho}_j$, and $\forall t \in \bbR$, $y(t) := \overline{\bY}(t)$, with a slight abuse of notation for $y$, which is a signal written in lowercase letter instead of capital bold letter.
We want to compute the Hessian of
\begin{equation}
    F(\tau, \rho)
    :=
    \int_0^T \left( y(\rho (t - \Lambda \tau)) - y(t) \right)^2 dt \enspace ,
\end{equation}
with respect to $\tau \in \bbR$ and $\rho \in \bbR$.
Let us define the function 
\begin{equation}
    f : (t, \tau, \rho) 
    \mapsto
    \left( y(\rho (t - \Lambda \tau)) - y(t) \right)^2 \enspace .
\end{equation}
We denote $H_F(\tau, \rho) \in \bbR^{2 \times 2}$ (resp. $H_f(\tau, \rho)(t) \in \bbR^{2 \times 2}$) the Hessian of $F$ (resp. of $f(t, \cdot, \cdot)$) with respect to $\tau$ and $\rho$.
Using Leibniz integral rule, we have
\begin{equation}
    H_F(\tau, \rho)
    =
    \int_0^T H_f(\tau, \rho)(t) dt \enspace ,
\end{equation}
where taking the integral of a matrix means integrating each element of the matrix individually.

Our goal is to compare how fast function $F$ varies when $\tau$ and $\rho$ vary around $0$ and $1$, respectively.
To do so, we consider the elements in the diagonal of the matrix $H_F(\tau, \rho) \in \bbR^{2 \times 2}$, that is 
\begin{equation}
    \int_0^T \frac{\delta^2 f}{\delta \tau^2} (t, \tau, \rho) dt
    \quad \text{and} \quad
    \int_0^T \frac{\delta^2 f}{\delta \rho^2} (t, \tau, \rho) dt \enspace .
\end{equation}
Simple calculations give
\begin{align}
    \frac{\delta^2 f}{\delta \tau^2} (t, \tau, \rho)
    =
    2 \Lambda^2 \rho^2 \Bigl[ y'^2(\rho (t - \Lambda \tau)) \Bigr.
    +
    \Bigl. y''(\rho (t - \Lambda \tau)) [y(\rho (t - \Lambda \tau)) - y(t)] \Bigr] \enspace ,
\end{align}
where $\Lambda^2$ is the square of $\Lambda$, and not the hyperparameter from section~\ref{sssec:Lambda_2}.
Furthermore,
\begin{align}
    \frac{\delta^2 f}{\delta \rho^2} (t, \tau, \rho)
    =
    2 (t - \Lambda \tau)^2 \Bigl[ y'^2(\rho (t - \Lambda \tau)) \Bigr.
    +
    \Bigl. y''(\rho (t - \Lambda \tau)) [y(\rho (t - \Lambda \tau)) - y(t)] \Bigr] \enspace .
\end{align}
Thus, at the point $(\tau, \rho) = (0, 1)$, these second-order derivatives simplify to:
\begin{equation}
    \frac{\delta^2 f}{\delta \tau^2} (t, 0, 1)
    =
    2 \Lambda^2 y'^2(t) \enspace ,
\end{equation}
and
\begin{equation}
    \frac{\delta^2 f}{\delta \rho^2} (t, 0, 1)
    =
    2 t^2 y'^2(t) \enspace .
\end{equation}

Consequently, the diagonal of the matrix $H_F(0, 1) \in \bbR^{2 \times 2}$ is, up to the factor $2$,
\begin{equation}
    \left( \Lambda^2 \! \int_0^T y'^2(t) dt, \int_0^T t^2 y'^2(t) dt \right) \enspace .
\end{equation}
So that the diagonal contains two times the same number, we must choose
\begin{equation}
    \Lambda
    =
    \sqrt{\frac{\int_0^T t^2 y'^2(t) dt}{\int_0^T y'^2(t) dt}} \enspace .
\end{equation}
Reusing index $j$, we obtain the desired quantity
\begin{equation}
    \Lambda_j = n \sqrt{\frac{\int_0^T t^2 (\nabla_t \overline{\bY}_j(t))^2 dt}{\int_0^T (\nabla_t \overline{\bY}_j(t))^2 dt}} \enspace,
\end{equation}
which concludes the proof.

\end{proof}

\section{Hyperparameter \texorpdfstring{$\Lambda^2$}{Lambda2}}
\label{appendix:parameters_scales_2}

\begin{figure}[!h]
    \centerline{\includegraphics[width=0.55\textwidth]{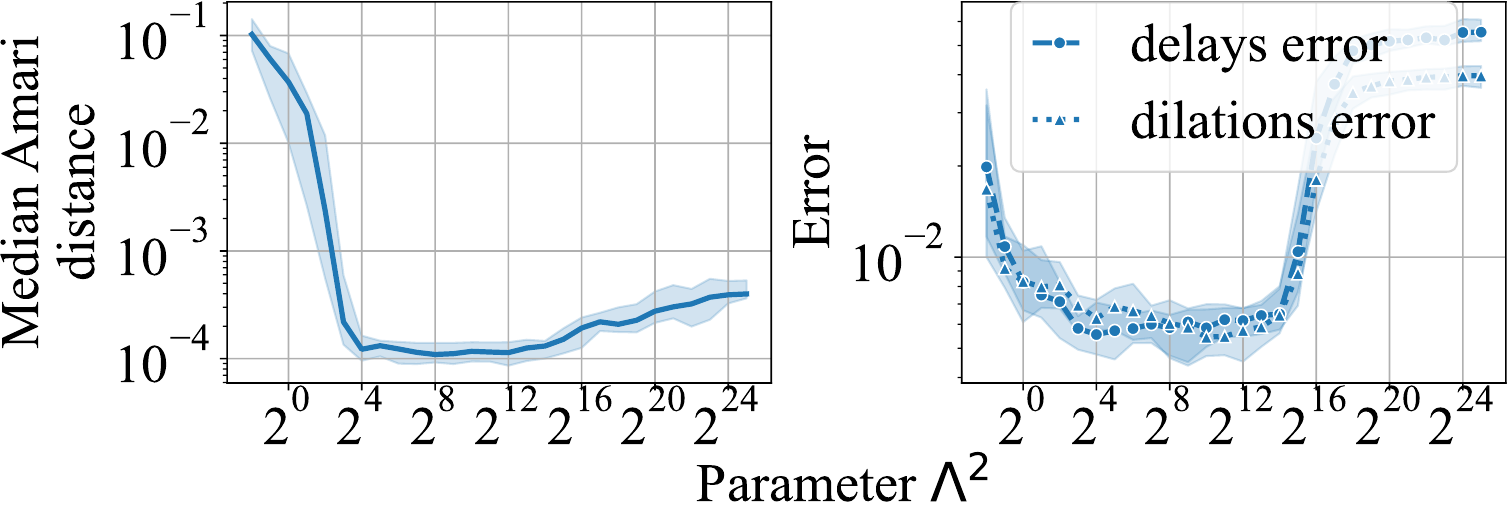}}
    \caption{
    \centering
    Amari distance (left) and time parameters errors (right) of our method as a function of the hyperparameter $\Lambda^2$. \added{The results indicate that $\Lambda^2$ can be chosen from a wide range of values within $[2^4, 2^{14}]$ without affecting performance, making careful tuning unnecessary.}}
    \label{fig:figures_wrt_W_scale}
\end{figure}

This section provides details on selecting the hyperparameter $\Lambda^2$, which controls the scale of delays and dilations relative to that of the unmixing matrices. Figure \ref{fig:figures_wrt_W_scale} illustrates that, with our synthetic data settings ($5$ views, $3$ sources, $600$ samples, $\tau_{\max}=0.05$, and $\rho_{\max}=1.15$), the Amari distance drops sharply until $\Lambda^2$ reaches $2^4$. Additionally, the errors on delays and dilations increase significantly when $\Lambda^2$ exceeds $2^{14}$. Therefore, in this context, $\Lambda^2$ should be selected within $[2^4, 2^{14}]$, a range that fortunately does not require high specificity. This selection causes MVICAD$^2$ to prioritize optimizing delays and dilations over unmixing matrices, especially during the initial iterations.

\newpage

\section{Initialization}
\label{appendix:initialization}

Algorithm \ref{algo:initialization} describes in detail the initialization algorithm.
It works as follows.
A single-view ICA algorithm, like Picard \cite{ablin2018faster}, is applied to each view $\bX^i$ to get the individual sources $\bS^i$.
Then, first-view sources $\bS^1$ are chosen as the reference and we iterate over the other sources $\bS^i$, $i \geq 2$.
For each source number $j \in [\![ p ]\!]$, we iterate over $k \in [\![ p ]\!]$ to check if $\bS^1_j$ and $\bS^i_k$ correspond to the same source.
For such a pair, we test several combinations of delay $\tau^i_k$ and dilation $\rho^i_k$, each one over a grid with $n_{\text{grid}}$ points (typically set to $10$).
Only the combination that produces the highest correlation between $\bS^1_j$ and $\overleftarrow{\cT}_{\tau^i_k, \rho^i_k}(\bS^i_k)$ is stored.
Finally, we use the Hungarian algorithm to recover the optimal pairs of sources $(j, k)$. 
In the end, this procedure arranges the individual sources to match the order of the first-view's sources and identifies delays and dilations that approximately align them.
\begin{algorithm}
\caption{Initialization}
\label{algo:initialization}
\begin{algorithmic}
    \STATE \textbf{Input:} observations $\cX = (\bX^1, \dots, \bX^m) \in \bbR^{m \times p \times n}$, number of grid points $n_{\text{grid}}$, maximum delay $\tau_{\max}$, and maximum dilation $\rho_{\max}$
    \STATE \textbf{Output:} $\cW, \btau$, and $\brho$
    \FOR {$i \gets 1$ to $m$}
        \STATE $W^i, \bS^i = \text{ICA}(\bX^i)$
    \ENDFOR
    \STATE $\btau_{\text{grid}} \gets$ grid of $n_{\text{grid}}$ equally spaced points of $[-\tau_{\max}, \tau_{\max}]$
    \STATE $\brho_{\text{grid}} \gets$ grid of $n_{\text{grid}}$ equally spaced points of $[1 / \rho_{\max}, \rho_{\max}]$
    \STATE $\btau^1 \gets$ vector of zeros, $\brho^1 \gets$ vector of ones
    \FOR {$i \gets 2$ to $m$}
        \STATE $C \gets 0^{p \times p}, T \gets 0^{p \times p}, P \gets 0^{p \times p}$
        \FOR {$j \gets 1$ to $p$}
            \STATE $\bs_{\text{ref}} \gets \bS^1_j$
            \FOR {$k \gets 1$ to $p$}
                \STATE $\bs_{\text{actual}} \gets \bS^i_k$
                \STATE $c_{\text{best}} \gets +\infty, \tau_{\text{best}} \gets \emptyset, \rho_{\text{best}} \gets \emptyset$
                \FOR {$\tau \gets$ first to last element of $\btau_{\text{grid}}$}
                    \FOR {$\rho \gets$ first to last element of $\brho_{\text{grid}}$}
                        \STATE $\bs_{\text{actual}} \gets \bs_{\text{actual}}$ delayed by $\tau$ and dilated by $\rho$
                        \STATE $c \gets$ cross-correlation score of $\bs_{\text{ref}}$ and $\bs_{\text{actual}}$
                        \IF {$c < c_{\text{best}}$}
                            \STATE $c_{\text{best}} \gets c, \tau_{\text{best}} \gets \tau, \rho_{\text{best}} \gets \rho$
                        \ENDIF
                    \ENDFOR
                \ENDFOR
                \STATE $C_{jk} \gets c_{\text{best}}, T_{jk} \gets \tau_{\text{best}}, P_{jk} \gets \rho_{\text{best}}$
            \ENDFOR
        \ENDFOR
        \STATE $\text{order}^i \gets \text{Hungarian}(C)$
        \STATE $W^i \gets W^i[\text{order}^i], \btau^i \gets T[:, \text{order}^i], \brho^i \gets P[:, \text{order}^i]$
    \ENDFOR
    \STATE $\cW \gets (W^1, \dots, W^m), \btau \gets (\btau^1, \dots, \btau^m), \brho \gets (\brho^1, \dots, \brho^m)$
\end{algorithmic}
\end{algorithm}

\newpage

\section{Main algorithm}\label{appendix:main_algorithm}

Algorithm \ref{algo:main} summarizes the entire pipeline.
It starts with the initialization in Algorithm \ref{algo:initialization} and proceeds through the steps outlined in Section II of the paper.
To speed up computations, we use Just-In-Time (JIT) compilation of the loss function and its gradient.
\begin{algorithm}
\caption{MVICAD\textsuperscript{2}}
\label{algo:main}
\begin{algorithmic}
    \STATE \textbf{Input:} observations $\cX = (\bX^1, \dots, \bX^m)$, maximum delay $\tau_{\max}$, maximum dilation $\rho_{\max}$, scale of the time parameters $\Lambda^2$
    \STATE \textbf{Output:} $\cY$, $\btau$, and $\brho$
    \STATE Initialize $\cW$, $\btau$, and $\brho$ with Algorithm \ref{algo:initialization} detailed in Appendix \ref{appendix:initialization}
    \FOR {$i \gets 1$ to $m$}
        \STATE $\bY^i \gets \overleftarrow{\cT}_{\btau^i, \brho^i} (W^i \bX^i)$
    \ENDFOR
    \STATE $\overline{\bY} \gets \frac{1}{m} \sum_{i=1}^m \bY^i$
    \STATE Compute $\Lambda^1 = (\Lambda_1, \ldots, \Lambda_p)^\top \in \bbR^p$
    \FOR {$i \gets 1$ to $m$}
        \STATE $\btau^i \gets \tau_{\text{scale}} \odot \btau^i$ \; where \; $\tau_{\text{scale}} \gets \frac{\Lambda^2}{\tau_{\max}} \Lambda^1 \in \bbR^p$
        \STATE $\brho^i \gets \rho_{\text{scale}} \brho^i$ \; where \; $\rho_{\text{scale}} \gets \frac{\Lambda^2}{\rho_{\max} - 1} \in \bbR$
    \ENDFOR
    \STATE Set the box constraints to $]-\infty, +\infty[$ for each element of $\cW$, $[-\tau_{\text{scale}} \tau_{\max}, \tau_{\text{scale}} \tau_{\max}]$ for each row of $\btau$, and $[\frac{1}{\rho_{\max}} \rho_{\text{scale}}, \rho_{\max} \rho_{\text{scale}}]$ for each element of $\brho$
    \STATE Perform JIT compilation of the loss function $\tilde{\tilde{\cL}}$ and its gradient with JAX \cite{jax2025}
    \STATE Use the L-BFGS-B algorithm \cite{byrd1995limited} to optimize $\tilde{\tilde{\cL}}$ with respect to $\cW$, $\btau$, and $\brho$, with the above-mentioned box constraints
    \FOR {$i \gets 1$ to $m$}
        \STATE $\bY^i \gets \overleftarrow{\cT}_{\btau^i, \brho^i} (W^i \bX^i)$
    \ENDFOR
    \STATE $\cY \gets (\bY^1, \dots, \bY^m)$
\end{algorithmic}
\end{algorithm}

\section{Synthetic experiment without delay or dilation}
\label{appendix:expe_without_delay_or_dilation}

\begin{figure}[h]
    \centerline{\includegraphics[width=0.55\textwidth]{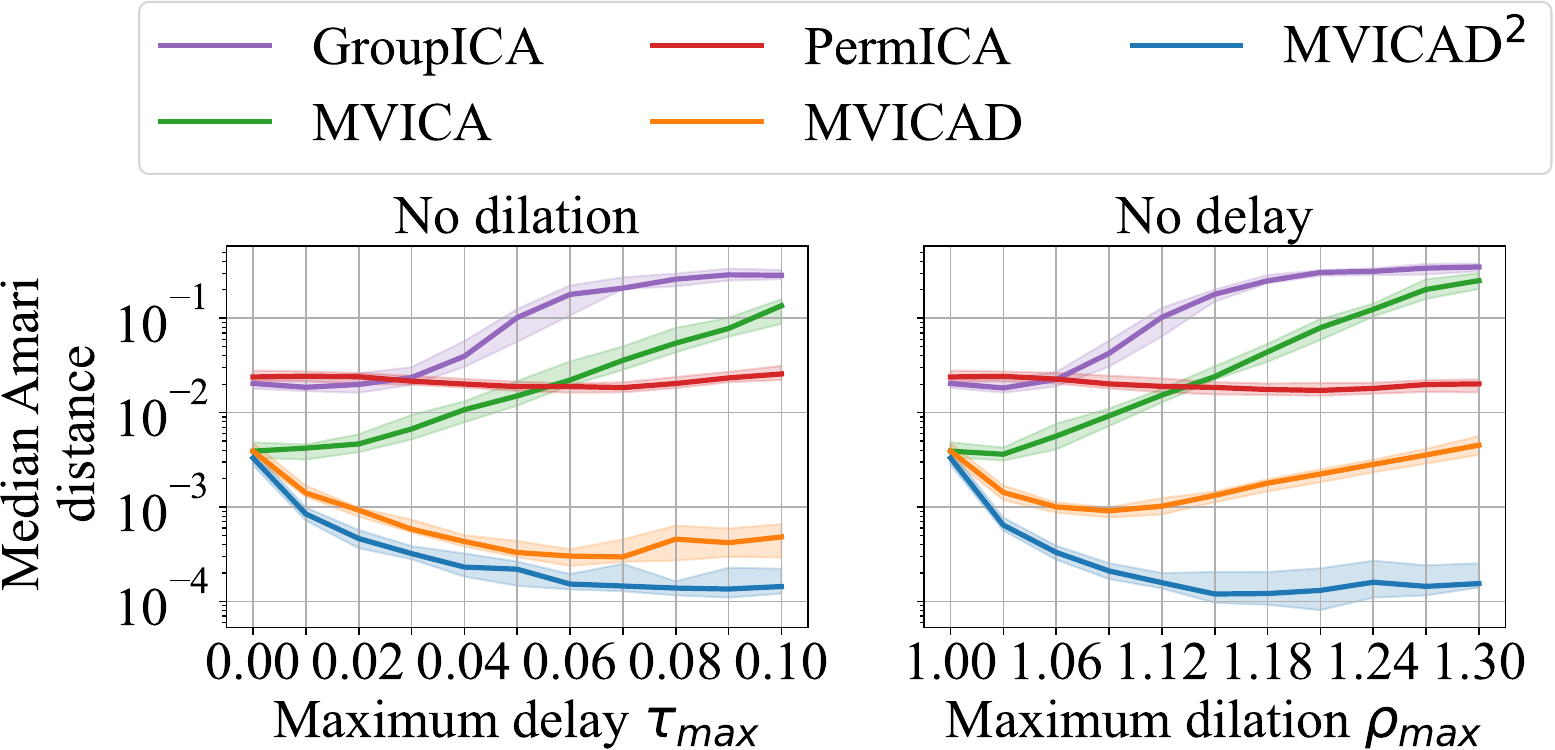}}
    \caption{
    \centering
    \added{Amari distance of multiple methods in two scenarios. Left: there is no dilation, and we vary the maximum delay $\tau_{\max}$. Right: there is no delay, and we vary the maximum dilation $\rho_{\max}$.
    We used $50$ different seeds and plotted the median of all seeds.}}
    \label{fig:amari_distance_wrt_max_delay_and_max_dilation_separately}
\end{figure}

\added{To further evaluate the robustness of our method in potentially adversarial conditions, we conducted an additional experiment using the same data generation protocol as in Figure~\ref{fig:amari_distance_wrt_max_dilation_and_max_shift}. In this new setting, we independently varied the strength of delays (resp. dilations) while removing dilations (resp. delays). The resulting median Amari distances are reported in Figure~\ref{fig:amari_distance_wrt_max_delay_and_max_dilation_separately}.}

\added{The results confirm expectations and support the validity of our approach. When there is no dilation (left panel), MVICAD$^2$ and MVICAD achieve the lowest Amari distance, with MVICAD$^2$ performing slightly better. This improvement likely stems from three factors: (1) MVICAD$^2$ estimates delays as continuous variables rather than as discrete integers, (2) it directly optimizes the full objective rather than a partial one, and (3) it employs a different optimization strategy.}

\added{When there is no delay (right panel), MVICAD$^2$ clearly outperforms all other methods, as it is the only one explicitly designed to account for dilations. This result highlights the importance of modeling dilations when present in the data.}

\added{Additionally, when there is no delay and no dilation (i.e. leftmost points in both panels), MVICA, MVICAD, and MVICAD$^2$ all converge to similar performance, as expected. This confirms that the extensions introduced in MVICAD and MVICAD$^2$ do not degrade performance when unnecessary.}

%% file: main_arxiv.bbl
\begin{thebibliography}{}

\bibitem[Ablin et~al., 2018]{ablin2018faster}
Ablin, P., Cardoso, J.-F., and Gramfort, A. (2018).
\newblock Faster independent component analysis by preconditioning with hessian approximations.
\newblock {\em IEEE Transactions on Signal Processing}, 66(15):4040--4049.

\bibitem[Alain and Woods, 1999]{alain1999age}
Alain, C. and Woods, D.~L. (1999).
\newblock Age-related changes in processing auditory stimuli during visual attention: evidence for deficits in inhibitory control and sensory memory.
\newblock {\em Psychology and aging}, 14(3):507.

\bibitem[Bell and Sejnowski, 1995]{bell1995information}
Bell, A.~J. and Sejnowski, T.~J. (1995).
\newblock An information-maximization approach to blind separation and blind deconvolution.
\newblock {\em Neural computation}, 7(6):1129--1159.

\bibitem[Bradbury et~al., 2018]{jax2025}
Bradbury, J., Frostig, R., Hawkins, P., Johnson, M.~J., Leary, C., Maclaurin, D., Necula, G., Paszke, A., VanderPlas, J., Wanderman-Milne, S., and Zhang, Q. (2018).
\newblock {JAX}: composable transformations of {Python+NumPy} programs.
\newblock GitHub repository.
\newblock Version 0.3.13, accessed August 2025.

\bibitem[Brookes et~al., 2011]{brookes2011investigating}
Brookes, M.~J., Woolrich, M., Luckhoo, H., Price, D., Hale, J.~R., Stephenson, M.~C., Barnes, G.~R., Smith, S.~M., and Morris, P.~G. (2011).
\newblock Investigating the electrophysiological basis of resting state networks using magnetoencephalography.
\newblock {\em Proceedings of the National Academy of Sciences}, 108(40):16783--16788.

\bibitem[Byrd et~al., 1995]{byrd1995limited}
Byrd, R.~H., Lu, P., Nocedal, J., and Zhu, C. (1995).
\newblock A limited memory algorithm for bound constrained optimization.
\newblock {\em SIAM Journal on scientific computing}, 16(5):1190--1208.

\bibitem[Calhoun et~al., 2009]{calhoun2009review}
Calhoun, V.~D., Liu, J., and Adal{\i}, T. (2009).
\newblock A review of group {ICA} for {fMRI} data and {ICA} for joint inference of imaging, genetic, and {ERP} data.
\newblock {\em Neuroimage}, 45(1):S163--S172.

\bibitem[Cantlon et~al., 2011]{cantlon2011cortical}
Cantlon, J.~F., Pinel, P., Dehaene, S., and Pelphrey, K.~A. (2011).
\newblock Cortical representations of symbols, objects, and faces are pruned back during early childhood.
\newblock {\em Cerebral cortex}, 21(1):191--199.

\bibitem[Chen et~al., 2015]{chen2015reduced}
Chen, P.-H.~C., Chen, J., Yeshurun, Y., Hasson, U., Haxby, J., and Ramadge, P.~J. (2015).
\newblock A reduced-dimension fmri shared response model.
\newblock {\em Advances in neural information processing systems}, 28:460--468.

\bibitem[Comon, 1994]{comon1994independent}
Comon, P. (1994).
\newblock Independent component analysis, a new concept?
\newblock {\em Signal processing}, 36(3):287--314.

\bibitem[Correa et~al., 2010]{correa2010canonical}
Correa, N.~M., Adali, T., Li, Y.-O., and Calhoun, V.~D. (2010).
\newblock Canonical correlation analysis for data fusion and group inferences.
\newblock {\em IEEE signal processing magazine}, 27(4):39--50.

\bibitem[Curran et~al., 2001]{curran2001effects}
Curran, T., Hills, A., Patterson, M.~B., and Strauss, M.~E. (2001).
\newblock Effects of aging on visuospatial attention: an erp study.
\newblock {\em Neuropsychologia}, 39(3):288--301.

\bibitem[Finnigan et~al., 2011]{finnigan2011erp}
Finnigan, S., O'Connell, R.~G., Cummins, T.~D., Broughton, M., and Robertson, I.~H. (2011).
\newblock Erp measures indicate both attention and working memory encoding decrements in aging.
\newblock {\em Psychophysiology}, 48(5):601--611.

\bibitem[Haxby et~al., 2011]{haxby2011common}
Haxby, J.~V., Guntupalli, J.~S., Connolly, A.~C., Halchenko, Y.~O., Conroy, B.~R., Gobbini, M.~I., Hanke, M., and Ramadge, P.~J. (2011).
\newblock A common, high-dimensional model of the representational space in human ventral temporal cortex.
\newblock {\em Neuron}, 72(2):404--416.

\bibitem[Heurtebise et~al., 2023]{heurtebise2023multiview}
Heurtebise, A., Ablin, P., and Gramfort, A. (2023).
\newblock Multiview independent component analysis with delays.
\newblock In {\em 2023 IEEE 33rd International Workshop on Machine Learning for Signal Processing (MLSP)}, pages 1--6. IEEE.

\bibitem[Hyvarinen, 1999]{hyvarinen1999fast}
Hyvarinen, A. (1999).
\newblock Fast and robust fixed-point algorithms for independent component analysis.
\newblock {\em IEEE transactions on Neural Networks}, 10(3):626--634.

\bibitem[Hyv{\"a}rinen and Oja, 2000]{hyvarinen2000independent}
Hyv{\"a}rinen, A. and Oja, E. (2000).
\newblock Independent component analysis: algorithms and applications.
\newblock {\em Neural networks}, 13(4-5):411--430.

\bibitem[Kamronn et~al., 2015]{kamronn2015multiview}
Kamronn, S., Poulsen, A.~T., and Hansen, L.~K. (2015).
\newblock Multiview bayesian correlated component analysis.
\newblock {\em Neural computation}, 27(10):2207--2230.

\bibitem[Kettenring, 1971]{kettenring1971canonical}
Kettenring, J.~R. (1971).
\newblock Canonical analysis of several sets of variables.
\newblock {\em Biometrika}, 58(3):433--451.

\bibitem[Kuhn, 1955]{kuhn1955hungarian}
Kuhn, H.~W. (1955).
\newblock The hungarian method for the assignment problem.
\newblock {\em Naval research logistics quarterly}, 2(1-2):83--97.

\bibitem[Liu and Nocedal, 1989]{liu1989limited}
Liu, D.~C. and Nocedal, J. (1989).
\newblock On the limited memory bfgs method for large scale optimization.
\newblock {\em Mathematical programming}, 45(1):503--528.

\bibitem[Makeig et~al., 2004]{makeig2004mining}
Makeig, S., Debener, S., Onton, J., and Delorme, A. (2004).
\newblock Mining event-related brain dynamics.
\newblock {\em Trends in cognitive sciences}, 8(5):204--210.

\bibitem[Moreau and Macchi, 1998]{moreau1998self}
Moreau, E. and Macchi, O. (1998).
\newblock Self-adaptive source separation. ii. comparison of the direct, feedback, and mixed linear network.
\newblock {\em IEEE transactions on signal processing}, 46(1):39--50.

\bibitem[Pichora-Fuller et~al., 2017]{pichora2017older}
Pichora-Fuller, M.~K., Alain, C., and Schneider, B.~A. (2017).
\newblock Older adults at the cocktail party.
\newblock {\em The auditory system at the cocktail party}, pages 227--259.

\bibitem[Price et~al., 2017]{price2017age}
Price, D., Tyler, L.~K., Neto~Henriques, R., Campbell, K.~L., Williams, N., Treder, M.~S., Taylor, J.~R., and Henson, R. (2017).
\newblock Age-related delay in visual and auditory evoked responses is mediated by white-and grey-matter differences.
\newblock {\em Nature communications}, 8(1):15671.

\bibitem[Richard et~al., 2020]{richard2020modeling}
Richard, H., Gresele, L., Hyvarinen, A., Thirion, B., Gramfort, A., and Ablin, P. (2020).
\newblock Modeling shared responses in neuroimaging studies through multiview ica.
\newblock {\em Advances in Neural Information Processing Systems}, 33:19149--19162.

\bibitem[Sabuncu et~al., 2010]{sabuncu2010function}
Sabuncu, M.~R., Singer, B.~D., Conroy, B., Bryan, R.~E., Ramadge, P.~J., and Haxby, J.~V. (2010).
\newblock Function-based intersubject alignment of human cortical anatomy.
\newblock {\em Cerebral cortex}, 20(1):130--140.

\bibitem[Schmolesky et~al., 2000]{schmolesky2000degradation}
Schmolesky, M.~T., Wang, Y., Pu, M., and Leventhal, A.~G. (2000).
\newblock Degradation of stimulus selectivity of visual cortical cells in senescent rhesus monkeys.
\newblock {\em Nature neuroscience}, 3(4):384--390.

\bibitem[Schneider et~al., 2010]{schneider2010effects}
Schneider, B.~A., Pichora-Fuller, K., and Daneman, M. (2010).
\newblock Effects of senescent changes in audition and cognition on spoken language comprehension.
\newblock {\em The aging auditory system}, pages 167--210.

\bibitem[Sun, 2013]{sun2013survey}
Sun, S. (2013).
\newblock A survey of multi-view machine learning.
\newblock {\em Neural computing and applications}, 23:2031--2038.

\bibitem[Taylor et~al., 2017]{taylor2017cambridge}
Taylor, J.~R., Williams, N., Cusack, R., Auer, T., Shafto, M.~A., Dixon, M., Tyler, L.~K., Henson, R.~N., et~al. (2017).
\newblock {The Cambridge Centre for Ageing and Neuroscience (Cam-CAN) data repository: Structural and functional MRI, MEG, and cognitive data from a cross-sectional adult lifespan sample}.
\newblock {\em neuroimage}, 144:262--269.

\bibitem[Tichavsky and Koldovsky, 2004]{tichavsky2004optimal}
Tichavsky, P. and Koldovsky, Z. (2004).
\newblock Optimal pairing of signal components separated by blind techniques.
\newblock {\em IEEE Signal Processing Letters}, 11(2):119--122.

\bibitem[Tsatsishvili et~al., 2015]{tsatsishvili2015combining}
Tsatsishvili, V., Cong, F., Toiviainen, P., and Ristaniemi, T. (2015).
\newblock Combining {PCA} and multiset {CCA} for dimension reduction when group {ICA} is applied to decompose naturalistic {fMRI} data.
\newblock In {\em 2015 International Joint Conference on Neural Networks (IJCNN)}, pages 1--6. IEEE.

\bibitem[Varoquaux~G and B, 2009]{varoquaux2009canica}
Varoquaux~G, Sadaghiani~S, P.~J. and B, T. (2009).
\newblock {CanICA: Model-based extraction of reproducible group-level ICA patterns from fMRI time series}.
\newblock In {\em {Medical Image Computing and Computer Aided Intervention}}, page~1, London, United Kingdom.

\bibitem[Walton, 2010]{walton2010timing}
Walton, J.~P. (2010).
\newblock Timing is everything: temporal processing deficits in the aged auditory brainstem.
\newblock {\em Hearing research}, 264(1-2):63--69.

\bibitem[Xu et~al., 2013]{xu2013survey}
Xu, C., Tao, D., and Xu, C. (2013).
\newblock A survey on multi-view learning.
\newblock {\em arXiv preprint arXiv:1304.5634}.

\bibitem[Zhao et~al., 2017]{zhao2017multi}
Zhao, J., Xie, X., Xu, X., and Sun, S. (2017).
\newblock Multi-view learning overview: Recent progress and new challenges.
\newblock {\em Information Fusion}, 38:43--54.

\end{thebibliography}
